\documentclass[twoside,11pt]{article}

\usepackage{jmlr2e}
\usepackage{comment} 
\usepackage[ruled]{algorithm2e} 
\usepackage{boondox-cal} 
\usepackage{setspace} 
\usepackage{amsmath} 
\usepackage{amssymb}
\usepackage{calc}
\usepackage{bm} 
\usepackage[T1]{fontenc}
\usepackage{framed,color}

\newcommand{\E}{{\mathbb{E}}}
\newcommand{\w}{{\bf w}}

\newtheorem{condition}{Condition}
\newtheorem{lem}{Lemma} 

\ShortHeadings{Optimal Weighted Random Forests}{Chen, Yu and Zhang}
\firstpageno{1}

\begin{document}

\title{Optimal weighted random forests}

\author{\name Xinyu Chen
	\email SA21204192@mail.ustc.edu.cn \\
       \addr School of Management\\
            University of Science and Technology of China\\
            Hefei, 230026, Anhui, China
       \AND
       \name Dalei Yu
       \email yudalei@126.com \\
       \addr School of Mathematics and Statistics\\
       Xi'an Jiaotong University, Xi'an, 710049, Shaan'xi\\
       China
       \AND
       \name Xinyu Zhang
       \email xinyu@amss.ac.cn \\
       \addr Academy of Mathematics and Systems Science\\
            Chinese Academy of Sciences\\
            Beijing, 100190, China\\
            School of Management\\
            University of Science and Technology of China\\
            Hefei, 230026, Anhui, China
       }

\editor{editor1 and editor2}

\maketitle

\begin{abstract}

The random forest (RF) algorithm  has become a very popular prediction method for its great flexibility and promising accuracy. In RF, it is conventional to put equal weights on all the base learners (trees) to aggregate their predictions. However, the predictive performances of different trees within the forest can be very different due to the randomization of the embedded bootstrap sampling and feature selection. In this paper, we focus on RF for regression and propose two optimal weighting algorithms, namely the 1 Step Optimal Weighted RF (1step-WRF$_\mathrm{opt}$) and 2 Steps Optimal Weighted RF (2steps-WRF$_\mathrm{opt}$), that combine the base learners through the weights determined by weight choice criteria. Under some regularity conditions, we show that these algorithms are asymptotically optimal in the sense that the resulting squared loss and risk are asymptotically identical to those of the infeasible but best possible model averaging estimator. Numerical studies conducted on real-world data sets indicate that these algorithms outperform the equal-weight forest and two other weighted RFs proposed in existing literature in most cases.
\end{abstract}


\section{Introduction}\label{section:introduction}
Random forest (RF) \citep{breiman2001random} is one of the most successful machine learning algorithms that scale with the volume of information while maintaining sufficient statistical efficiency \citep{Biau:Scornet:2016}.
Due to its great flexibility and promising accuracy, RF has been widely used in diverse areas of data analysis, including policy-making \citep{yoon2021forecasting, lin2021analyzing}, business analysis \citep{pallathadka2021applications, ghosh2022forecasting}, chemoinformatics \citep{svetnik2003random}, real-time human pose recognition  \citep{Shotton:etal:2011}, and so on. RF ensembles multiple decision trees grown on bootstrap samples and yields highly accurate predictions. In the conventional implementation of RF, it is customary and convenient to allocate equal weight to each decision tree.
Theoretically, the predictive performance varies from tree to tree due to the application of randomly selected sub-spaces of data and features. In other words, trees exhibit greater diversity owing to the injected randomness. An immediate question then arises: Is it always optimal to consider equal weights? In fact, there is sufficient evidence indicating that an averaging strategy with appropriately selected unequal weights may achieve  better performance than simple averaging (i.e., equal weighting) if individual learners exhibit non-identical strength~\citep{zhou2012ensemble,Peng:Yang:2022}.

To solve the problem mentioned above, some efforts have been made in the literature regarding weighted RFs. Specifically, Trees Weighting Random Forest (TWRF) introduced by~\citet{li2010trees} adopts the accuracy in the out-of-bag data as an index that measures the classification power of the tree and sets it as the weight. \citet{winham2013weighted} develop Weighted Random Forests (wRF), where the weights are determined based on tree-level prediction error. Based on wRF, \citet{xuan2018refined} put forward Refined Weighted Random Forests (RWRF) using all training data, including in-bag data and out-of-bag data. A novel weights formula is also developed in RWRF but cannot be manipulated into a regression pattern. \citet{pham2019cesaro} replace the regular average with a Ces\'aro average with theoretical analysis. However, these studies have predominantly focused on classification and less attention has been paid to the regression pattern (i.e., estimating the conditional expectation), although some mechanisms for classification can be transformed into corresponding regression patterns. In addition, none of the aforementioned studies have investigated the theoretical underpinnings regarding the optimality properties of their methods.

Recently, \cite{qiu2020mallows} propose a novel framework that averages the outputs of multiple machine learning algorithms by the weights determined from Mallows-type criteria. The authors further demonstrate that their framework can be applied to tree-type algorithms, employing regression tree, bagging regression tree and RF as base learners, respectively. Motivated by their work, we extend this approach by developing an asymptotically optimal weighting strategy for RF. Specifically, we treat the individual trees within the RF as base learners and employ Mallows-type criteria to obtain their respective weights.
Besides, to reduce computational burden, we further propose an accelerated algorithm that only requires two quadratic optimization tasks. Asymptotic optimality is established for both the original and accelerated weighted RF estimators. Extensive analyses on real-world data sets demonstrate that the proposed methods show promising performance over existing RFs.

The remaining part of the paper proceeds as follows: Section~\ref{section:Model and Problem Formulation} formulates the problem. Section~\ref{section:Mallows-type Averaging Random Forest} establishes our weighted RF algorithms and provides theoretical analysis. Section~\ref{section:Real Data Analysis} shows their promising performance on 12 real-world data sets from UCI Machine Learning Repository. Section~\ref{section:Conclusion} concludes.

\section{Model and Problem Formulation}\label{section:Model and Problem Formulation}

Let $\{x_{ij}\}$ be a set of $p$ predictors (or explanatory variables) and $y_i$ be a univariate response variable for $i=1,\ldots,n$ and $j=1,\ldots,p$.
Consider a data sample of
${\{y_i,\mathbf{x}_i\}}_{i=1}^{n}$, where $\mathbf{x}_i={(x_{i1},\ldots,x_{ip})}^{\top}$. The data generating process is as follows
$$
y_i=\mu_i+e_i,
$$
where $e_i$ is the random error with $\mathbb{E}(e_i|\mathbf{x}_i)=0$ and $\mathbb{E}(e_{i}^{2}|\mathbf{x}_i)=\sigma_{i}^{2}$, and $\mu_i=\mathbb{E}(y_i|\mathbf{x}_i)$. So heteroscedasticity is allowed here.

Given a predictor vector $\mathbf{x}_i$, the corresponding prediction for $y_i$ by a tree (or base learner, BL) in the construction of RF can be written as follows
$$
\widehat{y}_i=\mathbf{P}_{\mathrm{BL}}^{\top}(\mathbf{x}_{i},\mathbf{X},\mathbf{y},\mathbf{\mathcal{B}},\mathbf{\Theta}) \mathbf{y},
$$
where $\mathbf{y}={(y_{1},\ldots,y_{n})}^{\top}$ is the vector of the response variable and  $\mathbf{X}={(\mathbf{x}_{1},\ldots,\mathbf{x}_{n})}^{\top}$ is the matrix of predictors. The variables $\mathbf{\mathcal{B}}$ and $\mathbf{\Theta}$ are
not considered explicitly but play implicit roles in injecting randomness. First, each tree is fit to an independent bootstrap sample from
the original data. The randomization involved in bootstrap sampling makes up $\mathbf{\mathcal{B}}$. Second, The randomization used to split variables and to cut points at each node furnishes the component
of $\mathbf{\Theta}$. The nature and
dimension of  $\mathbf{\mathcal{B}}$ and $\mathbf{\Theta}$ depend on tree construction.

Let us assume that we have drawn $M_n$ bootstrap data sets of size $n$ and grown $M_n$ trees on their bootstrapped data, where $M_n$ can grow with $n$ or remains fixed. Take the $m^{\mathrm{th }}$  tree for example.
Dropping an instance $(y_0,\mathbf{x}_0)$ down this base learner and end up with a specific tree leaf $l$ with $n_l$ observations $\mathcal{D}_l=\{(y_{i_1},\mathbf{x}_{i_1}),\cdots,(y_{i_{n_l}},\mathbf{x}_{i_{n_l}})\}$. Assume that the number of occurrences of instance $(y_i,\mathbf{x}_i)$ in this tree is $h_i$ for all $i$ because of the bootstrap sampling procedure. Then $\mathbf{P}_{\mathrm{BL}}(\mathbf{x}_0, \mathbf{X},\mathbf{y},\mathbf{\mathcal{B}},\mathbf{\Theta})$ for this tree is a sparse vector, with elements of ${h_i}/{n_l}$ and zero otherwise, corresponding to the counterparts in $\mathcal{D}_l$ between $(y_0,\mathbf{x}_0)$ and ${\{y_i,\mathbf{x}_i\}}_{i=1}^{n}$. Specifically, the $i^{\mathrm{th}}$ element of $\mathbf{P}_{\mathrm{BL}}(\mathbf{x}_0, \mathbf{X},\mathbf{y},\mathbf{\mathcal{B}},\mathbf{\Theta})$ is ${h_i}/{n_l}$, if $(y_i,\mathbf{x}_i) \in \mathcal{D}_l$, and zero otherwise. Elements of $\mathbf{P}_{\mathrm{BL}}(\mathbf{x}_0, \mathbf{X},\mathbf{y},\mathbf{\mathcal{B}},\mathbf{\Theta})$ are weights put on elements of $\mathbf{y}$ to make a prediction for $y_0$.

By randomly selecting sub-spaces of data and features, trees in RFs are given more randomness than trees without these randomization techniques. Specifically, bootstrapped data are used to generate trees rather than the original training data. In addition, instead of using all of the predictors before splitting at each node, we draw $q$ $(q<p)$ variables from a total of $p$ variables. We have
$h_i\equiv 1$ for all $i \in \{1,\ldots,n\}$, and $\mathbf{P}_{\mathrm{BL}}(\mathbf{x}_0, \mathbf{X},\mathbf{y},\mathbf{\mathcal{B}},\mathbf{\Theta})$ will contain fewer zero elements, if without the bootstrap procedure.

The prediction for $y_i$ by the $m^{ \mathrm{th }}$ tree (or the $m^{ \mathrm{th }}$ base learner) within the forest obeys the following relationship
$$
\widehat{y}_{i}^{(m)}=\mathbf{P}_{\mathrm{BL}{(m)}}^{\top}(\mathbf{x}_{i},\mathbf{X},\mathbf{y},{\mathbf{\mathcal{B}}_{(m)}},{\mathbf{\Theta}_{(m)}}) \mathbf{y},
$$
where $\widehat{y}_{i}^{(m)}$ is the prediction for $y_i$ by the $m^{\mathrm{th }}$ tree, and $\mathbf{P}_{\mathrm{BL}{(m)}}(\mathbf{x}_{i},\mathbf{X},\mathbf{y},{\mathbf{\mathcal{B}}_{(m)}},{\mathbf{\Theta}_{(m)}})$ is the vector $\mathbf{P}_{\mathrm{BL}}(\mathbf{x}_{i},\mathbf{X},\mathbf{y},\mathbf{\mathcal{B}},\mathbf{\Theta})$ related to the $m^{\mathrm{th }}$ tree.
The final output of the forest is integrated by
$$
\widehat{y}_{i}(\mathbf{w})=\sum_{m=1}^{M_n} w_{(m)} \widehat{y}_{i}^{(m)},
$$
where $w_{(m)}$ is the weight put on the $m^{\mathrm{th}}$ tree.

Our goal is to determine appropriate weights to improve prediction accuracy of RF, given a predictor vector $\mathbf{x}$. Clearly, the conventional RF has $w_{(m)} \equiv 1/M_n$ for $m=1,\ldots,M_n$.

\section{Mallows-type Weighted RFs}\label{section:Mallows-type Averaging Random Forest}

Let $\mathbf{P}_{\mathrm{BL}{(m)}}$ be an $n \times n$ matrix, of which the $i^{\mathrm{th }}$ row is $\mathbf{P}_{\mathrm{BL}{(m)}}^{\top}(\mathbf{x}_{i},\mathbf{X},\mathbf{y},{\mathbf{\mathcal{B}}_{(m)}},{\mathbf{\Theta}_{(m)}})$.
Let $\mathbf{P}(\mathbf{w})=\sum_{m=1}^{M_n} w_{(m)} \mathbf{P}_{\mathrm{BL}{(m)}}$, and $\widehat{\mathbf{y}}(\mathbf{w}) = \sum_{m=1}^{M_n} w_{(m)} \widehat{\mathbf{y}}^{(m)}$ with $\widehat{\mathbf{y}}^{(m)} = {(\widehat{y}_{1}^{(m)},\ldots,\widehat{y}_{n}^{(m)})}^{\top}$.  Define the following averaged squared error function
\begin{equation}
L_{n}(\mathbf{w}) \equiv\|\widehat{\mathbf{y}}(\mathbf{w})-\boldsymbol{\mu}\|^{2},
\label{averaged squared error  function}
\end{equation}
which measures the sum of squared biases between the true $\boldsymbol{\mu}=\mathbb{E}(\mathbf{y}|\mathbf{X})$ and its model averaging estimate $\widehat{\mathbf{y}}(\mathbf{w})$.
Let $\ell_{(m)}$ be the number of leaves in the $m^{\mathrm{ th}}$ tree ,
$n_{(m), l}$ be the number of observations in the $l^{\mathrm{th}}$ leaf of the $m^{\mathrm{ th}}$ tree,
and $R_{n}(\mathbf{w})=\mathbb{E}\left\{L_{n}(\mathbf{w})| \mathbf{X}\right\}$.
We will suggest criteria to obtain weights based on $R_{n}(\mathbf{w})$.

\subsection{Mallows-type Weight Choice Criteria}
Considering the choice of weights, we use the solution obtained by minimizing the following Mallows-type criterion (\ref{C2}) with the restriction of $\mathbf{w} \in \mathcal{H} \equiv\left\{\mathbf{w} \in[0,1]^{M_n}: \sum_{m=1}^{M_n} w_{(m)}=1\right\}$

\begin{equation}
C_n(\mathbf{w})=\|\mathbf{y}-\mathbf{P}(\mathbf{w}) \mathbf{y}\|^{2}+2 \sum_{i=1}^{n} e_{i}^{2} P_{ii}(\mathbf{w}),\label{C2}
\end{equation}
where $P_{i i}(\mathbf{w})$ is the $i^{\mathrm{th}}$ diagonal term in $\mathbf{P}(\mathbf{w})$, and $\mathbf{e}={(e_1,\ldots,e_n)}^{\top}$ is the true error term vector.

This criterion is originally proposed by~\citet{zhao2016model} for considering linear models.  In the context of linear models, $\E\{C_n(\mathbf{w})\}$ equals  the expected predictive squared error up to a constant. \citet{zhao2016model} further show that the criterion is asymptotically optimal in the context considered therein. However, $e_i$'s are unobservable terms in practice. So they further consider the following feasible
criterion, replacing the true error terms with averaged residuals

\begin{equation}
C_{n}^{\prime}(\mathbf{w})=\|\mathbf{y}-\mathbf{P}(\mathbf{w}) \mathbf{y}\|^{2}+2 \sum_{i=1}^{n} \hat{e}_{i}(\mathbf{w})^{2} P_{ii}(\mathbf{w}),\label{criterion2'}
\end{equation}
where
$$\hat{\mathbf{e}}(\mathbf{w})={\left[\hat{e}_{1}(\mathbf{w}),\ldots,\hat{e}_{n}(\mathbf{w})\right]}^{\top}=\sum_{m=1}^{M_n} w_{(m)} \hat{\mathbf{e}}^{(m)}=\{\mathbf{I}-\mathbf{P}(\mathbf{w})\}\mathbf{y},$$
$\hat{\mathbf{e}}^{(m)}$ is the residual vector for the $m^{\mathrm{th}}$ candidate model, and $\mathbf{I}$ is the identity matrix. This feasible criterion also accommodates heteroscedasticity. Besides, it relies on all candidate models to estimate the true error vector, which avoids placing too much confidence on a single model. Similar criterion has also been considered in \citet{qiu2020mallows}.

We apply criterion (\ref{criterion2'}) to  determine $\w$ in $\widehat{\mathbf{y}}(\mathbf{w})$.
Criterion (\ref{criterion2'}) comprises of two terms. The first term measures the fitting error of the weighted RF in the training data, by computing the residual sum of squares. The second term penalizes the complexity of the trees in the forest.
For each $m \in \{1,\ldots,M_n\}$, $P_{\mathrm{BL}(m)i i}$ denotes the $i^{\mathrm{th}}$ diagonal term in $\mathbf{P}_{\mathrm{BL}(m)}$. As explained in Section \ref{section:Model and Problem Formulation}, $P_{\mathrm{BL}(m)i i}$ is the proportion of the  $i^{\mathrm{th}}$ observation to the total number of samples in the leaf that includes the $i^{\mathrm{th}}$ observation.
Thus, for each $m \in \{1,\ldots,M_n\}$ and $i \in \{1,\ldots,n\}$, the larger the value of  $P_{\mathrm{BL}(m)i i}$, the smaller the gap between $y_i$ and $\widehat{y}_{i}^{(m)}$. In the extreme case where a tree is so deep that the leaf node containing the $i^{\mathrm{th}}$ observation is pure, $P_{\mathrm{BL}(m)i i}$ equals 1, and $\widehat{y}_{i}^{(m)}$ equals $y_i$. Essentially, this tree has low prediction error within the training sample, but may exhibit poor generalization performance when applied to new data.
To mitigate the contribution of overfitted trees in the ensembled output,
this algorithm assigns a lower weight to these trees, thereby decreasing the second term.

From another perspective, assuming homoscedasticity, the weighted residual terms $\hat{e}_{i}(\mathbf{w})^{2}$ for all $i=1,\ldots,n$ share the same value, and can be moved outside the summation. Then, the summation part $\sum_{i=1}^{n} P_{ii}(\mathbf{w})=\sum_{m=1}^{M_n} w_{(m)} \sum_{i=1}^{n} P_{\mathrm{B L}(m)ii}=\sum_{m=1}^{M_n} w_{(m)} \ell_{(m)}$ represents the weighted number of leaf nodes of all trees. The regularized objective for minimizing in Extreme Gradient Boosting (XGBoost) algorithm, proposed by \citet{chen2016xgboost}, also contains a penalty term that penalizes the number of leaves in the tree. In light of this, both the weighted RF with weights obtained by minimizing criterion (\ref{criterion2'}) and XGBoost employ the number of tree leaves in a tree to measure its complexity. Intuitively, criterion (\ref{criterion2'}) tends to assign higher weights to trees that exhibit lower prediction errors on the training sample and show better generalization performance outside the training sample.

It is clear that criterion (\ref{criterion2'}) is a cubic function of $\mathbf{w}$, whose optimization is substantially more time-consuming than that of quadratic programming.
To expedite the process, we further suggest an accelerated algorithm that estimates ${\mathbf{e}}$ using a vector that is independent of $\mathbf{w}$.
The accelerated algorithm consists of two steps, where the first step involves calculating the estimated error terms, and the second step involves substituting the vector obtained in the first step for the true error terms in criterion (\ref{C2}). In specific, consider the following intermediate criterion,
\begin{equation}
C_{n}^{0}(\mathbf{w})=\|\mathbf{y}-\mathbf{P}(\mathbf{w}) \mathbf{y}\|^{2}+2 \sum_{i=1}^{n} \hat{\sigma}^{2} P_{ii}(\mathbf{w}),\label{criterion1}
\end{equation}
where $\hat{\sigma}^{2}=\|\mathbf{y}-\mathbf{P}(\mathbf{w}_0) \mathbf{y}\|^{2} / n$, and $\mathbf{w}_0$ is a vector with all elements equal $1/M_n$.
Solve this quadratic programming task over $\mathbf{w} \in \mathcal{H}$ and get a solution $\mathbf{w}^{*}$. Then, calculate the residual vector by
$$\tilde{\mathbf{e}}={(\tilde{e}_{1},\ldots,\tilde{e}_{n})}^{\top}=\{\mathbf{I}-\mathbf{P}(\mathbf{w}^{*})\}\mathbf{y}.$$
Next, consider the following criterion
\begin{equation}
C_{n}^{\prime \prime}(\mathbf{w})=\|\mathbf{y}-\mathbf{P}(\mathbf{w}) \mathbf{y}\|^{2}+2 \sum_{i=1}^{n} \tilde{e}_{i}^{2} P_{ii}(\mathbf{w})\label{criterion2''}.
\end{equation}
Both (\ref{criterion1}) and (\ref{criterion2''}) are quadratic functions of $\w$, while criterion (\ref{criterion2'}) is a cubic function. Many contemporary software packages, such as quadprog in R or MATLAB, can effectively handle quadratic programming problems. In fact, from the real data analysis conducted in Section \ref{section:Real Data Analysis}, it is observed that the time required to solve two quadratic programming problems is notably lower compared to that required to solve a more intricate nonlinear programming problem of higher order. Please see Table \ref{Time Consumption Comparisons} for more details.

We refer to the RF with tree-level weights derived from optimizing criterion (\ref{criterion2'}) as 1 Step Optimal Weighted RF (1step-WRF$_\mathrm{opt}$), and the RF with weights of trees obtained by optimizing criterion (\ref{criterion2''}) as 2 Steps Optimal Weighted RF (2steps-WRF$_\mathrm{opt}$). Their details are provided in Algorithms \ref{Algorithm OWRF} and \ref{Algorithm 2step-OWRF}, respectively.

Before providing the theoretical support of the proposed algorithms,
we introduce a tree-type algorithm that aims to construct a tree whose structure is independent of the output values of the learning sample. Such configuration has also been imposed in the theoretical analysis of \citet{geurts2006extremely} and \cite{biau2012analysis}. Moreover, this theoretical framework is referred to as ``honesty'' in the field of causal inference \citep{athey2016recursive} and is essential for further theoretical analysis. We term this tree as Split-Unsupervised Tree (SUT) in contrast to Classification and Regression Tree (CART) whose split criterion relies on response information.
Under this setup, the vector $\mathbf{P}_{\mathrm{BL}}(\mathbf{x}_{i},\mathbf{X},\mathbf{y},\mathbf{\mathcal{B}},\mathbf{\Theta})$ reduces to $\mathbf{P}_{\mathrm{BL}}(\mathbf{x}_{i},\mathbf{X},\mathbf{\mathcal{B}},\mathbf{\Theta})$. The details of the SUT and CART algorithms are provided in Appendix A.

\begin{algorithm}[h]
	\setstretch{1.3}
	\caption{1step-WRF$_\mathrm{opt}$}
	\label{Algorithm OWRF}
	\LinesNumbered
	\KwIn {(1) The training data set $\mathcal{D}={\{y_i,\mathbf{x}_i\}}_{i=1}^{n}$
		(2) The number of trees in random forest $M_n$}
	\KwOut {Weight vector $\widehat{\mathbf{w}} \in \mathcal{H}$}
	
	\For{m = 1 to $M_n$}{
		Draw a bootstrap data set $\mathcal{D}_{(m)}$ of size $n$ from the training data set $\mathcal{D}$;
		
		Grow a random-forest tree $\hat{f}_{(m)}$ to the bootstrapped data $\mathcal{D}_{(m)}$, by recursively repeating the following steps for each terminal node of
		the tree, until the minimum node size $n_{\min}$ is reached \tcp*{$n_{\min}, q$ are hyper parameters}
		
		\qquad i. Select $q$ variables at random from the $p$ variables;
		
		\qquad ii. Pick the best variable/ split-point among the $q$;
		
		\qquad  iii. Split the node into two daughter nodes.
		
		\For{i = 1 to n}{
			Drop $\mathbf{x}_i$ down the tree $m$ and get $\mathbf{P}_{\mathrm{BL}{(m)}}(\mathbf{x}_{i},\mathbf{X},\mathbf{y},{\mathbf{\mathcal{B}}}_{(m)},{\mathbf{\Theta}}_{(m)})$ if CART trees or $\mathbf{P}_{\mathrm{BL}{(m)}}(\mathbf{x}_{i},\mathbf{X},{\mathbf{\mathcal{B}}}_{(m)},{\mathbf{\Theta}}_{(m)})$ if SUT trees;
		}
		
		$\mathbf{P}_{\mathrm{BL}{(m)}} \leftarrow {\{{\mathbf{P}_{\mathrm{BL}{(m)}}(\mathbf{x}_1,\mathbf{X},\mathbf{y},{\mathbf{\mathcal{B}}}_{(m)},{\mathbf{\Theta}}_{(m)})},\ldots,{\mathbf{P}_{\mathrm{BL}{(m)}}(\mathbf{x}_n,\mathbf{X},\mathbf{y},{\mathbf{\mathcal{B}}}_{(m)},{\mathbf{\Theta}}_{(m)})}\}}^{\top}$ if CART trees\\
		
		or $\mathbf{P}_{\mathrm{BL}{(m)}} \leftarrow {{\{\mathbf{P}_{\mathrm{BL}{(m)}}(\mathbf{x}_1,\mathbf{X},{\mathbf{\mathcal{B}}}_{(m)},{\mathbf{\Theta}}_{(m)})},\ldots,{\mathbf{P}_{\mathrm{BL}{(m)}}(\mathbf{x}_n,\mathbf{X},{\mathbf{\mathcal{B}}}_{(m)},{\mathbf{\Theta}}_{(m)})}\}}^{\top}$ if SUT trees;
	}

	Solve the convex optimization problem:\qquad \qquad \qquad
	\qquad \qquad \qquad
	 $\widehat{\mathbf{w}} ={(\widehat w_{(1)},\ldots,\widehat w_{(M_n)})}^{\top}\leftarrow \underset{\mathbf{w} \in \mathcal{H}}{\arg \min } C_{n}^{\prime}(\mathbf{w})$
	  with  $\mathbf{P}(\mathbf{w})=\sum_{m=1}^{M_n} w_{(m)} \mathbf{P}_{\mathrm{BL}{(m)}}$.	
\end{algorithm}

\begin{algorithm}[h]
	\setstretch{1.3}
	\caption{2steps-WRF$_\mathrm{opt}$}
	\label{Algorithm 2step-OWRF}
	\LinesNumbered
	\KwIn {(1) The training data set $\mathcal{D}={\{y_i,\mathbf{x}_i\}}_{i=1}^{n}$
		(2) The number of trees in random forest $M_n$}
	\KwOut {Weight vector $\widetilde{\mathbf{w}} \in \mathcal{H}$}
	
	\For{m = 1 to $M_n$}{
		Draw a bootstrap data set $\mathcal{D}_{(m)}$ of size $n$ from the training data set $\mathcal{D}$;
		
		Grow a random-forest tree $\hat{f}_{(m)}$ to the bootstrapped data $\mathcal{D}_{(m)}$, by recursively repeating the following steps for each terminal node of
		the tree, until the minimum node size $n_{\min}$ is reached \tcp*{$n_{\min}, q$ are hyper parameters}
		
		\qquad i. Select $q$ variables at random from the $p$ variables;
		
		\qquad ii. Pick the best variable/ split-point among the $q$;
		
		\qquad  iii. Split the node into two daughter nodes.
		
		\For{i = 1 to n}{
			Drop $\mathbf{x}_i$ down the tree $m$ and get $\mathbf{P}_{\mathrm{BL}{(m)}}(\mathbf{x}_{i},\mathbf{X},\mathbf{y},{\mathbf{\mathcal{B}}}_{(m)},{\mathbf{\Theta}}_{(m)})$ if CART trees or $\mathbf{P}_{\mathrm{BL}{(m)}}(\mathbf{x}_{i},\mathbf{X},{\mathbf{\mathcal{B}}}_{(m)},{\mathbf{\Theta}}_{(m)})$ if SUT trees;
		}
		
		$\mathbf{P}_{\mathrm{BL}{(m)}} \leftarrow {\{{\mathbf{P}_{\mathrm{BL}{(m)}}(\mathbf{x}_1,\mathbf{X},\mathbf{y},{\mathbf{\mathcal{B}}}_{(m)},{\mathbf{\Theta}}_{(m)})},\ldots,{\mathbf{P}_{\mathrm{BL}{(m)}}(\mathbf{x}_n,\mathbf{X},\mathbf{y},{\mathbf{\mathcal{B}}}_{(m)},{\mathbf{\Theta}}_{(m)})}\}}^{\top}$ if CART trees\\
		
		or $\mathbf{P}_{\mathrm{BL}{(m)}} \leftarrow {\{{\mathbf{P}_{\mathrm{BL}{(m)}}(\mathbf{x}_1,\mathbf{X},{\mathbf{\mathcal{B}}}_{(m)},{\mathbf{\Theta}}_{(m)})},\ldots,{\mathbf{P}_{\mathrm{BL}{(m)}}(\mathbf{x}_n,\mathbf{X},{\mathbf{\mathcal{B}}}_{(m)},{\mathbf{\Theta}}_{(m)})}\}}^{\top}$ if SUT trees;
	}

	Solve the quadratic programming  problem:\qquad \qquad \qquad
	\qquad \qquad \qquad
	$\mathbf{w}^{*}={\left( w_{(1)}^{*},\ldots,w_{(M_n)}^{*}\right)}^{\top} \leftarrow \underset{\mathbf{w} \in \mathcal{H}}{\arg \min } C_{n}^{0}(\mathbf{w})$
	with  $\mathbf{P}(\mathbf{w})=\sum_{m=1}^{M_n} w_{(m)} \mathbf{P}_{\mathrm{BL}{(m)}}$;
	
	$\tilde{\mathbf{e}} \leftarrow \{\mathbf{I}-\mathbf{P}(\mathbf{w}^{*})\} \mathbf{y} $ with $\mathbf{P}(\mathbf{w}^{*})=\sum_{m=1}^{M_n} w_{(m)}^{*} \mathbf{P}_{\mathrm{BL}{(m)}}$;
	
	Solve the quadratic programming  problem:\qquad \qquad \qquad
	\qquad \qquad \qquad
	$\widetilde{\mathbf{w}} ={\left(\widetilde w_{(1)},\ldots,\widetilde w_{(M_n)}\right)}^{\top}\leftarrow \underset{\mathbf{w} \in \mathcal{H}}{\arg \min } C_{n}^{\prime \prime}(\mathbf{w})$
	with  $\mathbf{P}(\mathbf{w})=\sum_{m=1}^{M_n} w_{(m)} \mathbf{P}_{\mathrm{BL}{(m)}}$.	
\end{algorithm}

\subsection{Asymptotic Optimality}
In this section, we will establish the asymptotic optimality of the 1step-WRF$_\mathrm{opt}$ estimator and 2steps-WRF$_\mathrm{opt}$ estimator with SUT trees.
Denote the selected weight vectors from $C_{n}^{\prime}(\mathbf{w})$ and $C_{n}^{\prime \prime}(\mathbf{w})$ by
$$
 \widehat{\mathbf{w}}=\underset{\mathbf{w} \in \mathcal{H}}{\arg \min } C_{n}^{\prime}(\mathbf{w}) \text{ and }
 \widetilde{\mathbf{w}}=\underset{\mathbf{w} \in \mathcal{H}}{\arg \min } C_{n}^{\prime \prime}(\mathbf{w}),
$$
respectively.
Let $\xi_{n}=\inf_{\mathbf{w} \in \mathcal{H}} R_{n}(\mathbf{w})$. The following theorems establish the asymptotic optimality of the 1step-WRF$_\mathrm{opt}$ estimator and 2steps-WRF$_\mathrm{opt}$ estimator, respectively. We will list and discuss technical conditions required for proofs of Theorems \ref{theorem OWRF} and 2 in Appendix C.1.

\begin{theorem}[Asymptotic Optimality for 1step-WRF$_\mathrm{opt}$]\label{theorem OWRF}
	Assume Conditions \ref{condition 3} - \ref{condition 5} in Appendix C.1 hold. Then, as $n\rightarrow \infty$,
	\begin{equation}
	\frac{L_{n}\left(\widehat{\mathbf{w}}\right)}{\inf_{\mathbf{w} \in \mathcal{H}} L_{n}(\mathbf{w})} \stackrel{p}{\rightarrow} 1.\label{asymptotic optimality:Ln'}
	\end{equation}
If, in addition, $\left\{L_n(\widehat{\mathbf{w}})-\xi_n\right\}\xi_{n}^{-1}$ is bounded above by a positive constant, almost surely, then
\begin{equation}
\frac{R_{n}(\widehat{\mathbf{w}})}{\inf_{\mathbf{w} \in \mathcal{H}} R_{n}(\mathbf{w})}  \rightarrow 1.\label{risk}
\end{equation}
\end{theorem}

\begin{theorem}[Asymptotic Optimality for 2steps-WRF$_\mathrm{opt}$]\label{theorem 2step-OWRF}
	Assume Conditions \ref{condition 3} - \ref{condition extra} in Appendix C.1 hold. Then, as $n\rightarrow \infty$,
	\begin{equation}
	\frac{L_{n}\left(\widetilde{\mathbf{w}}\right)}{\inf_{\mathbf{w} \in \mathcal{H}} L_{n}(\mathbf{w})} \stackrel{p}{\rightarrow} 1.\label{asymptotic optimality:Ln''}
	\end{equation}
If, in addition, $\left\{L_n(\widetilde{\mathbf{w}})-\xi_n\right\}\xi_{n}^{-1}$ is  bounded above by a positive constant, almost surely, then
	\begin{equation}
	\frac{R_{n}(\widetilde{\mathbf{w}})}{\inf_{\mathbf{w} \in \mathcal{H}} R_{n}(\mathbf{w})}  \rightarrow 1.\label{risk''}
	\end{equation}	
\end{theorem}

Results obtained in (\ref{asymptotic optimality:Ln'}) and  (\ref{asymptotic optimality:Ln''}) regard the asymptotic optimality in the sense of achieving the lowest possible squared loss, while (\ref{risk}) and (\ref{risk''}) yield asymptotic optimality in the sense of achieving the lowest possible squared risk.
Proofs of Theorems \ref{theorem OWRF} and \ref{theorem 2step-OWRF} are presented in Appendix C.3 and C.4.

\section{Real Data Analysis}\label{section:Real Data Analysis}
To assess the prediction performance of different weighted RFs in practical situations, we used 12 data sets from the UCI data repository for machine learning. Appendix B features a demonstration of two competitors, namely wRF and CRF. The details of the 12 data sets are listed in Table \ref{data_Description}.

\begin{table}[htbp]
	\centering
	\begin{tabular}{ccrr}
		\hline
		Data set & Abbreviation  & Attributes  & Samples\\
		\hline
		Boston Housing& BH&13&506\\
		Servo & Servo &4 &167\\

		Auto-mpg&AM& 8&398\\
		Concrete Compressive Strength &CCS&9 &1030\\
		Airfoil Self-Noise&ASN&5&1503\\
		Combined Cycle Power Plant&CCPP&4&9568\\
		Concrete Slump Test&CST&7&103\\
		Energy Efficiency&EE&8&768\\
		Parkinsons Telemonitoring&PT&20&5875\\
		QSAR aquatic toxicity&QSAR&8&546\\
		Synchronous Machine&SM&4&557\\
		Yacht Hydrodynamics& YH&6&308\\
		\hline
	\end{tabular}
	\caption{Summary of 12 Data Sets}\label{data_Description}
\end{table}	

For the sake of brevity, in the following, we will refer to each data set by its abbreviation. We randomly partitioned each data set into training data, testing data and validation data, in the ratio of $0.5 : 0.3 : 0.2$. The training data was used to construct trees and to calculate weights, and the test data was used to evaluate the predictive performance of different algorithms. The validation data was employed to select tuning parameters, such as the exponent in the expression for calculating weights in the wRF, and probability sequence in the SUT algorithm.

In this section, the number of trees $M_n$ was set to $100$. Before each split,
the dimension of random feature sub-space $q$ was set to $\lceil p/3 \rceil$, which is the default value in the regression mode of the R package randomForest. We set the minimum leaf size $n_{\min}$ to $\sqrt{n}$ in CART trees and $5$ in SUT trees, in order to control the depth of trees.
We also tried other values of $M_n$ and $n_{\min}$, and the patterns of the performance remain unchanged in general.

For each strategy, the number of replication was set to $D=1000$ and the forecasting performance was accessed by the following two criteria:
$$
\mathrm{MSFE}=\frac{1}{D\times n_{\text{test}}} \sum_{d=1}^{D} \sum_{i=1}^{n_{\text{test}}}(y_i-\widehat{y}_{i,d})^2  \text{ and } \text { MAFE }=\frac{1}{D\times n_{\text{test}}} \sum_{d=1}^{D} \sum_{i=1}^{n_{\text{test}}}\left|y_i-\widehat{y}_{i,d} \right|,
$$
where $n_{\text{test}}$ is the size of testing data, and $\widehat{y}_{i,d}$ is the forecast for $y_i$ in the $d^{\mathrm{ th}}$ repetition. MSFE and MAFE are abbreviations of ``Mean Squared Forecast Error'' and ``Mean Absolute Forecast Error'', respectively.
Next, we will exhibit the results of different weighting techniques on RFs with CART trees and RFs with SUT trees, respectively.

\subsection{RFs with CART Trees}
Tables \ref{real data Test Error MSE} and \ref{real data Test Error MAE} exhibit the risks of RFs with CART trees calculated by MSFE and MAFE, respectively. Each row in the tables presents the risks of different strategies, sorted in ascending order, with the corresponding values displayed in parentheses.

Regarding MSFE, the 1step-WRF$_\mathrm{opt}$ or 2steps-WRF$_\mathrm{opt}$ estimator manifests the best performance in 10 out of 12 data sets, whereas it exhibits the best performance in 9 out of 12 data sets in terms of MAFE. It is observed that the wRF becomes the best method in some data sets. Of all cases considered, the CRF is found to never be the best method. 
It is also noticeable that the 2steps-WRF$_\mathrm{opt}$ is superior to the 1step-WRF$_\mathrm{opt}$ in most cases, albeit with minor differences.

Table \ref{Time Consumption Comparisons} compares the time consumption of the 2steps-WRF$_\mathrm{opt}$ and 1step-WRF$_\mathrm{opt}$ algorithms for a single run, averaged over $D$ repetitions, with the ratio of the latter to the former in the fourth column. Apparently, the 2steps-WRF$_\mathrm{opt}$ can accelerate optimization by tens or hundreds of times when compared to the 1step-WRF$_\mathrm{opt}$, given that solving quadratic optimization is considerably faster than solving higher-order nonlinear optimization task.

\begin{table}[htbp]
	\centering
	\begin{tabular}{crrrrr}
		\hline
		\text{Data set} & \multicolumn{1}{c}{RF}&\multicolumn{1}{c}{2steps-WRF$_\mathrm{opt}$}  &\multicolumn{1}{c}{1step-WRF$_\mathrm{opt}$}&\multicolumn{1}{c}{wRF}&\multicolumn{1}{c}{CRF}\\
		\hline
		BH&$15.484^{(5)}$ &$13.958^{(1)}$&$14.038^{(2)}$ &$14.517^{(3)}$& $14.664^{(4)}$\\
		
		Servo&$1.610 ^{(5)}$& $0.825^{(1)}$&$ 0.836^{(2)}$& $  0.860  ^{(3)}$& $1.169^{(4)}$\\%
		
		AM&$9.709^{(5)}$&$ 9.272^{(3)}$ &$  9.407 ^{(4)}$& $9.077  ^{(1)} $& $9.078^{(2)}$\\%
		
		CCS &$60.460^{(5)}$& $50.004^{(1)}$&$50.048 ^{(2)}$& $52.868 ^{(3)}$& $54.065^{(4)}$\\
		
		ASN&$20.022^{(5)}$&  $14.572^{(1)}$&$14.575 ^{(2)}$& $15.611 ^{(3)}$& $17.054^{(4)}$\\
		
		CCPP&$18.016^{(5)}$& $16.065^{(2)}$&$ 16.062^{(1)}$& $ 16.237^{(3)}$& $ 16.556^{(4)}$\\
		
		CST&$26.421^{(5)}$&$19.877^{(1)}$& $ 20.074 ^{(2)}$& $21.500 ^{(3)}$& $22.534^{(4)}$\\%
		
		EE&$4.332 ^{(5)}$&$3.643^{(2)}$&  $  3.642 ^{(1)}$& $ 3.964 ^{(3)}$& $ 4.087^{(4)}$\\%
		
		PT&$14.641 ^{(5)}$&$8.653^{(2)}$&  $ 8.649  ^{(1)} $& $9.099 ^{(3)}$& $10.819^{(4)}$\\%
		
		QSAR&$1.436 ^{(5)}$&  $1.423^{(3)}$&$ 1.434  ^{(4)}$& $1.417 ^{(1)}$& $ 1.420^{(2)}$\\
		
		SM($\times10^{-4} $)& $6.981^{(5)}$&$3.403^{(1)}$ &$ 3.404 ^{(2)}$& $ 4.342^{(3)}$& $  5.214^{(4)}$\\
		
		YH&$35.442 ^{(5)}$&$3.727^{(1)}$&  $3.735^{(2)} $& $  5.603 ^{(3)}$& $13.422^{(4)}$\\
		\hline
	\end{tabular}
	\caption{Test Error Comparisons by MSFE for Different Forests with CART Trees}\label{real data Test Error MSE}
\end{table}

\begin{table}[htbp]
	\centering
	\begin{tabular}{crrrrr}
		\hline
		\text{Data set} & \multicolumn{1}{c}{RF}&\multicolumn{1}{c}{2steps-WRF$_\mathrm{opt}$}  &\multicolumn{1}{c}{1step-WRF$_\mathrm{opt}$}&\multicolumn{1}{c}{wRF}&\multicolumn{1}{c}{CRF}\\
		\hline
		BH&$2.608^{(5)}$&$ 2.536 ^{(1)}$&$2.549^{(3)}$&$ 2.539^{(2)}$&$ 2.562^{(4)}$\\
		
		Servo&$ 0.900^{(5)}$&$ 0.550^{(2)}$&$ 0.550 ^{(3)}$&$0.535 ^{(1)}$&$0.754^{(4)}$\\
		
		AM&$ 2.199 ^{(5)}$&$2.164^{(3)}$&$ 2.182^{(4)}$&$ 2.141 ^{(1)}$&$2.144^{(2)}$\\
		
		CCS&$ 6.092^{(5)}$&$ 5.500^{(1)}$&$ 5.503^{(2)}$&$ 5.668^{(3)}$&$ 5.751^{(4)}$\\
		
		ASN&$ 3.607^{(5)}$&$ 3.013^{(1)}$&$ 3.013^{(2)}$&$ 3.121^{(3)}$&$ 3.295^{(4)}$\\
		
		CCPP&$ 3.243^{(5)}$&$ 3.058^{(2)}$&$ 3.058^{(1)}$&$ 3.075^{(3)}$&$ 3.108^{(4)}$\\
		
		CST&$ 4.023^{(5)}$&$ 3.425^{(1)}$&$ 3.445^{(2)}$&$ 3.568^{(3)}$&$ 3.676^{(4)}$\\
		
		EE&$ 1.563^{(5)}$&$ 1.349^{(2)}$&$ 1.349 ^{(1)}$&$1.423^{(3)}$&$ 1.487^{(4)}$\\
		
		PT &$2.933^{(5)}$&$ 2.201^{(2)}$&$ 2.201^{(1)}$&$ 2.241^{(3)}$&$ 2.493^{(4)}$\\
		
		QSAR&$ 0.892^{(4)}$&$ 0.888^{(3)}$&$ 0.892^{(5)}$&$ 0.885 ^{(1)}$&$0.887^{(2)}$\\
		
		SM($\times10^{-2} $)&$ 2.044^{(5)}$&$ 1.374^{(1)}$&$ 1.375^{(2)}$&$ 1.551^{(3)}$&$ 1.746^{(4)}$\\
		
		YH&$ 3.877 ^{(5)}$&$1.182^{(1)}$&$ 1.182^{(2)}$&$ 1.358^{(3)}$&$ 2.329^{(4)}$\\
		
		\hline
	\end{tabular}
	\caption{Test Error Comparisons by MAFE for Different Forests with CART Trees}\label{real data Test Error MAE}
	
\end{table}

\begin{table}[htbp]
	\centering
	\begin{tabular}{crrr}
		\hline
		\text{Data set} & 2steps-WRF$_\mathrm{opt}$  &1step-WRF$_\mathrm{opt}$& Ratio\\
		\hline
		BH&$0.065$&$3.898$&$60.371$\\
		
		Servo&$0.072$&$1.347$&$18.778$\\%
		
		AM&$0.060$&$2.894$&$48.368$\\%
		
		CCS &$0.081$&$6.822$&$83.982$\\
		
		ASN&$0.094$&$6.468$&$68.840$\\
		
		CCPP&$0.566$&$2.785$&$4.916$\\
		
		CST&$0.075$&$2.061$&$27.630$\\%
		
		EE&$0.072$&$3.498$&$48.304$\\%
		
		PT&$0.291$&$40.445$&$138.967$\\%
		
		QSAR&$0.069$&$3.148$&$45.431$\\
		
		SM&$0.060$&$1.409$&$23.327$\\
		
		YH&$0.065$&$2.631$&$40.546$\\
		\hline
	\end{tabular}
	\caption{Time Consumption Comparisons (Unit: seconds)}\label{Time Consumption Comparisons}
	
\end{table}

\begin{figure}[htbp]
	\centering
	\includegraphics[width=\textwidth - 0.5in]{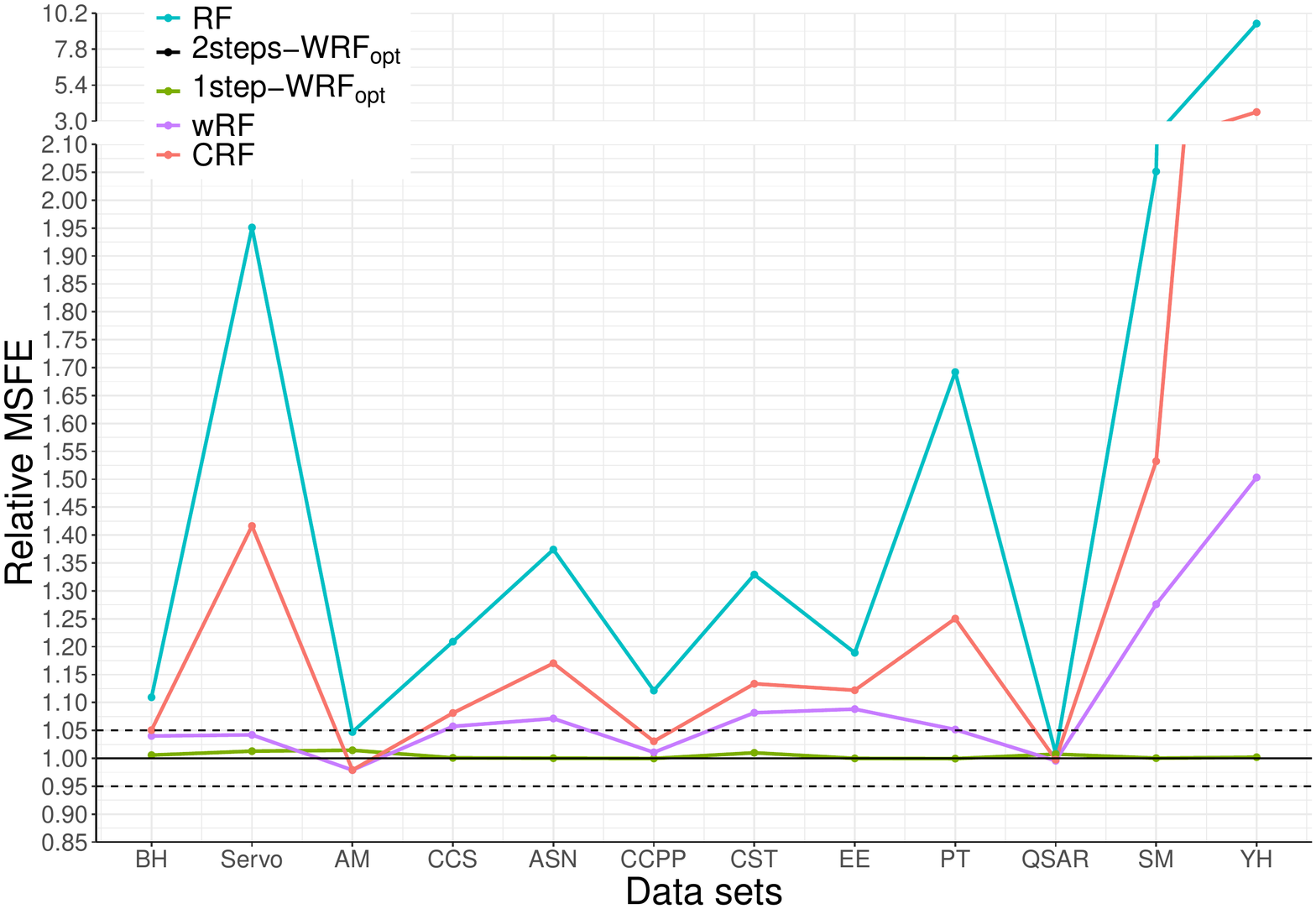}
	\caption{Relative MSFE for Different Forests with CART Trees}\label{real data Relative MSE}
\end{figure}

\begin{figure}[htbp]
	\centering
	\includegraphics[width=\textwidth - 0.5in]{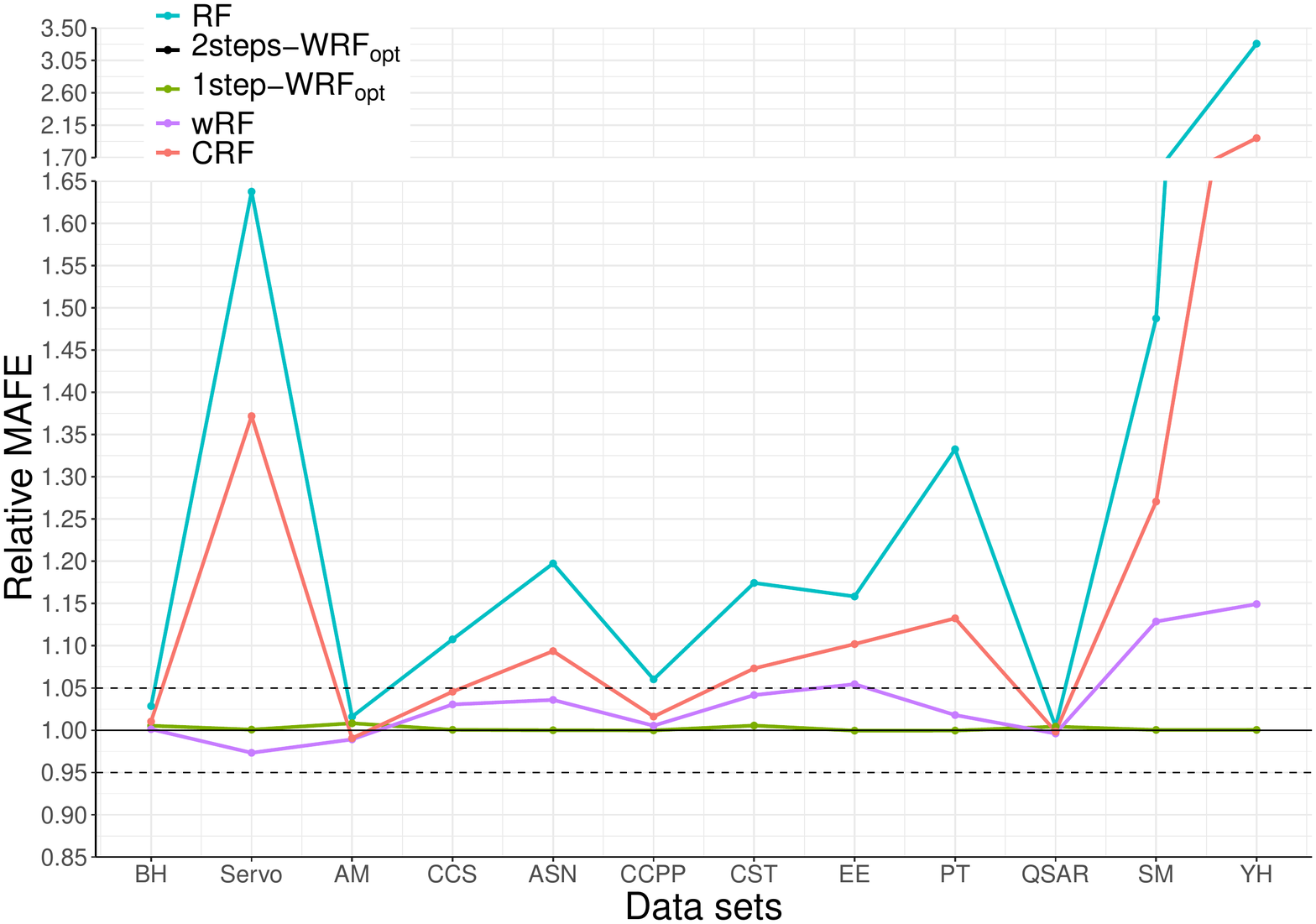}
	\caption{Relative MAFE for Different Forests with CART Trees}\label{real data Relative MAE}
\end{figure}

To further highlight their abilities in predictive accuracy,
we assessed relative risks based on the 2steps-WRF$_\mathrm{opt}$. More specifically, we divided the risks of the RF, 1step-WRF$_\mathrm{opt}$, wRF and CRF by the benchmark 2steps-WRF$_\mathrm{opt}$.
In the following, we assert that a relative risk is not essential if it falls in the interval of $(0.95, 1.05)$, while it is essential if it is lower than 0.95 or higher than 1.05.
The relative MSFE and MAFE of each method on 12 data sets are reported in Figures \ref{real data Relative MSE} and  \ref{real data Relative MAE}, respectively. The results are depicted by blue, black, green, purple and red lines, respectively, for the RF, 2steps-WRF$_\mathrm{opt}$, 1step-WRF$_\mathrm{opt}$, wRF, and CRF.

Some findings are worth mentioning in Figure \ref{real data Relative MSE}. First, the improvement of the WRF$_\mathrm{opt}$ (including the 1step-WRF$_\mathrm{opt}$ and 2steps-WRF$_\mathrm{opt}$) over the conventional RF is essential in 10 out of 12 data sets.
What stands out in the figure is that the relative MSFEs of others with respect to the benchmark
are conspicuously large in the YH data set. This spells the great success of our WRF$_\mathrm{opt}$ methods in practice.
More importantly, the WRF$_\mathrm{opt}$ outperforms competitors essentially in 7 out of 12 data sets, while none of the competitors dominate the benchmark essentially in all cases, underscoring the robustness of the WRF$_\mathrm{opt}$.

Figure \ref{real data Relative MAE} remains the similar qualitative results, albeit with less notable power of the WRF$_\mathrm{opt}$ than Figure \ref{real data Relative MSE}. Specifically, the WRF$_\mathrm{opt}$ shows essential improvement over the conventional RF in 9 out of 12 data sets, and dominates all competitors essentially in 3 out of 12 data sets. These proportions are relatively lower than those in Figure \ref{real data Relative MSE}. But none of the competitors surpass the benchmark essentially in all cases, which is consistent with Figure \ref{real data Relative MSE}.

Note that the wRF algorithm requires tuning a parameter outside of the training set, whereas the WRF$_\mathrm{opt}$ and CRF do not.
For the fairness of the comparison, all three weighted RFs should use identical tree models built in the same training data set.
As a result, the WRF$_\mathrm{opt}$ and CRF will use more training samples if not compared with the wRF, which may contribute to their superior predictive capability.
Combining all the findings together, we can conclude that the proposed WRF$_\mathrm{opt}$  yields more accurate predictions compared to the conventional RF and other existing weighted RFs in most cases.

\subsection{RFs with SUT Trees}
Similar to the previous scenario, Tables \ref{real data Test Error MSE (SUT)} and \ref{real data Test Error MAE (SUT)} display the risks of RFs with SUT trees computed by the MSFE and MAFE. The 1step-WRF$_\mathrm{opt}$ or 2steps-WRF$_\mathrm{opt}$ estimator consistently outperforms the conventional RFs and the two competitors in terms of MSFE,  while performing best in 11 out of 12 data sets in terms of MAFE. Additionally, the gaps between the 1step-WRF$_\mathrm{opt}$ and 2steps-WRF$_\mathrm{opt}$ are relatively small, akin to random forests with CART trees.

\begin{table}[htbp]
	\centering
	\begin{tabular}{crrrrr}
		\hline
		\text{Data set} & \multicolumn{1}{c}{RF}&\multicolumn{1}{c}{2steps-WRF$_\mathrm{opt}$}  &\multicolumn{1}{c}{1step-WRF$_\mathrm{opt}$}&\multicolumn{1}{c}{wRF}&\multicolumn{1}{c}{CRF}\\
		\hline
		BH &$ 38.213^{(5)}$&$  24.516 ^{(1)}$&$ 24.522 ^{(2)}$&$ 28.797 ^{(3)}$&$ 29.974^{(4)}$ \\
		
		Servo &$ 1.604 ^{(5)}$&$ 0.964 ^{(1)}$&$ 0.968 ^{(2)}$&$ 0.998 ^{(3)}$&$ 1.232^{(4)}$ \\
		
		AM &$ 13.952 ^{(5)}$&$ 9.632 ^{(1)}$&$ 9.642 ^{(2)}$&$ 11.508 ^{(3)}$&$ 12.148^{(4)}$ \\
		
		CCS &$ 149.276 ^{(5)}$&$ 119.471 ^{(1)}$&$ 119.505 ^{(2)}$&$ 136.243 ^{(4)}$&$ 132.435^{(3)}$ \\
		
		ASN &$ 36.391^{(5)}$&$ 33.465 ^{(1)}$&$ 33.472 ^{(2)}$&$ 35.675 ^{(4)}$&$ 35.419^{(3)}$ \\
		
		CCPP &$ 50.329 ^{(5)}$&$ 36.619 ^{(2)}$&$ 36.613 ^{(1)}$&$ 39.145 ^{(4)}$&$ 38.600^{(3)}$ \\
		
		CST &$ 42.899 ^{(5)}$&$ 25.933 ^{(2)}$&$ 25.928 ^{(1)}$&$ 32.806 ^{(3)}$&$ 36.783^{(4)}$ \\
		
		EE &$ 17.768 ^{(5)}$&$ 5.140 ^{(2)}$&$ 5.140 ^{(1)}$&$ 6.193 ^{(3)}$&$ 8.026 ^{(4)}$\\
		
		PT &$ 98.864 ^{(5)}$&$ 89.299 ^{(1)}$&$ 89.325 ^{(2)}$&$ 97.924 ^{(4)}$&$ 95.321 ^{(3)}$\\
		
		QSAR &$ 1.737 ^{(5)}$&$ 1.657 ^{(1)}$&$ 1.661 ^{(2)}$&$ 1.680 ^{(4)}$&$ 1.668^{(3)}$ \\
		
		SM($\times10^{-4} $) &$ 20.963 ^{(5)}$&$ 0.212 ^{(1)}$&$ 0.212 ^{(2)}$&$ 0.251 ^{(3)}$&$ 4.460 ^{(4)}$\\
		
		YH &$ 33.241 ^{(5)}$&$ 2.433 ^{(2)}$&$ 2.431 ^{(1)}$&$ 3.171 ^{(3)}$&$ 8.152 ^{(4)}$\\
		\hline
	\end{tabular}
	\caption{Test Error Comparisons by MSFE for Different Forests with SUT Trees}\label{real data Test Error MSE (SUT)}
\end{table}

\begin{figure}[htbp]
	\centering
	\includegraphics[width=\textwidth - 0.5in]{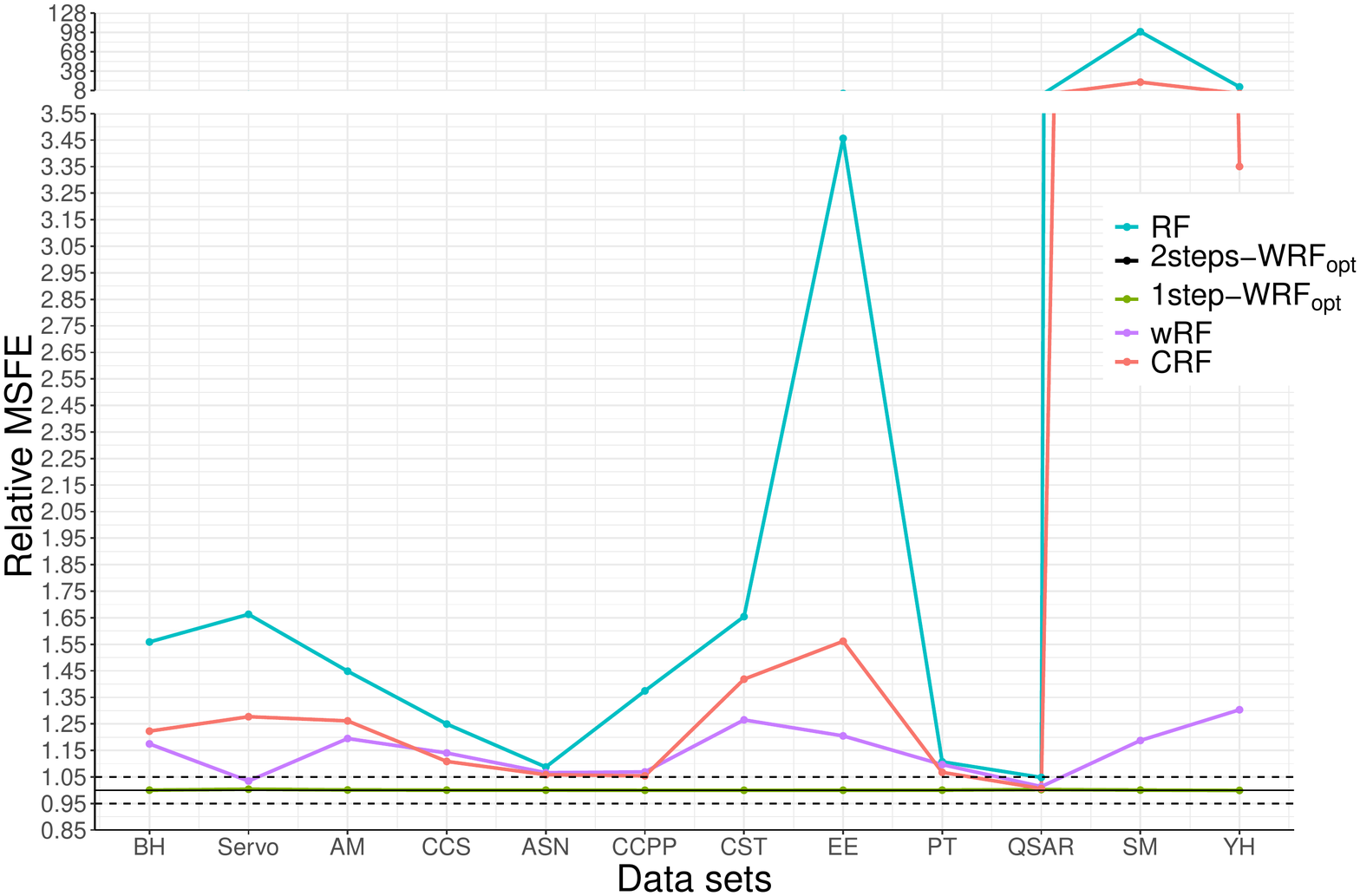}
	\caption{Relative MSFE for Different Forests with SUT Trees}\label{real data Relative MSE (SUT)}
\end{figure}

\begin{figure}[htbp]
	\centering
	\includegraphics[width=\textwidth - 0.5in]{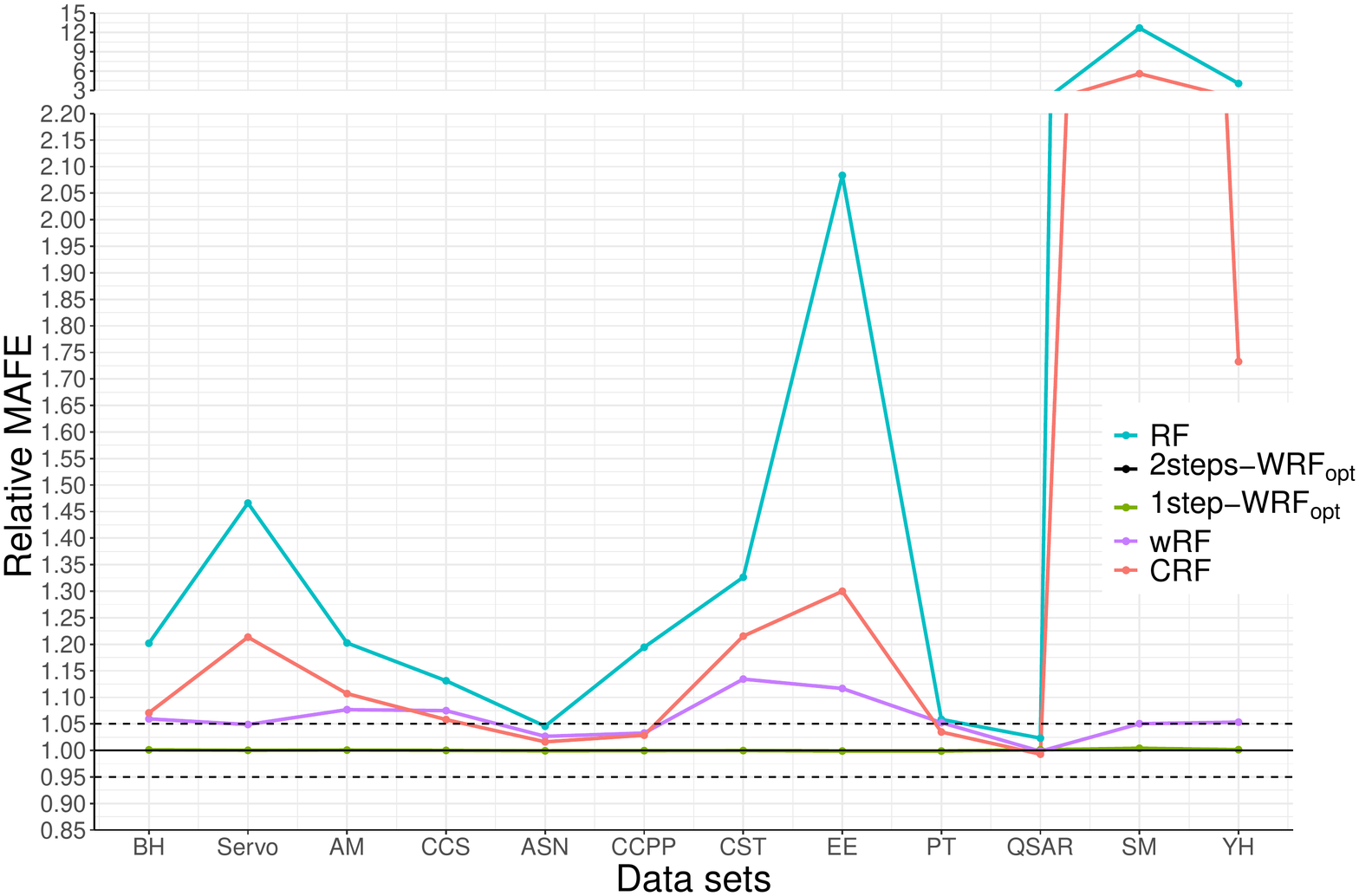}
	\caption{Relative MAFE for Different Forests with SUT Trees}\label{real data Relative MAE (SUT)}
\end{figure}

\begin{table}[htbp]
	\centering
	\begin{tabular}{crrrrr}
		\hline
		\text{Data set} & \multicolumn{1}{c}{RF}&\multicolumn{1}{c}{2steps-WRF$_\mathrm{opt}$}  &\multicolumn{1}{c}{1step-WRF$_\mathrm{opt}$}&\multicolumn{1}{c}{wRF}&\multicolumn{1}{c}{CRF}\\
		\hline
		BH &$ 3.759^{(5)}$&$ 3.128 ^{(1)}$&$ 3.131 ^{(2)}$&$ 3.314 ^{(3)}$&$ 3.349 ^{(4)}$\\
		
		Servo &$ 0.847 ^{(5)}$&$ 0.578 ^{(1)}$&$ 0.578 ^{(2)}$&$ 0.606 ^{(3)}$&$ 0.701 ^{(4)}$\\
		
		AM &$ 2.671 ^{(5)}$&$ 2.221 ^{(1)}$&$ 2.222 ^{(2)}$&$ 2.392 ^{(3)}$&$ 2.459 ^{(4)}$\\
		
		CCS &$ 9.914 ^{(5)}$&$ 8.764 ^{(1)}$&$ 8.764 ^{(2)}$&$ 9.421 ^{(4)}$&$ 9.273 ^{(3)}$\\
		
		ASN &$ 4.903 ^{(5)}$&$ 4.689 ^{(2)}$&$ 4.685 ^{(1)}$&$ 4.813 ^{(4)}$&$ 4.765^{(3)}$ \\
		
		CCPP &$ 5.805 ^{(5)}$&$ 4.860 ^{(2)}$&$ 4.858 ^{(1)}$&$ 5.019 ^{(4)}$&$ 4.999 ^{(3)}$\\
		
		CST &$ 5.209 ^{(5)}$&$ 3.927 ^{(2)}$&$ 3.926 ^{(1)}$&$ 4.455 ^{(3)}$&$ 4.773 ^{(4)}$\\
		
		EE &$ 3.356 ^{(5)}$&$ 1.611 ^{(2)}$&$ 1.609 ^{(1)}$&$ 1.799 ^{(3)}$&$ 2.094^{(4)}$ \\
		
		PT &$ 7.995 ^{(5)}$&$ 7.553 ^{(2)}$&$ 7.544 ^{(1)}$&$ 7.948 ^{(4)}$&$ 7.815^{(3)}$ \\
		
		QSAR &$ 1.001 ^{(5)}$&$ 0.979 ^{(3)}$&$ 0.980 ^{(4)}$&$ 0.977 ^{(2)}$&$ 0.972^{(1)}$ \\
		
		SM($\times10^{-2} $) &$ 3.674 ^{(5)}$&$ 0.290 ^{(1)}$&$ 0.291 ^{(2)}$&$ 0.304 ^{(3)}$&$ 1.628 ^{(4)}$\\
		
		YH &$ 3.508^{(5)}$&$ 0.856 ^{(1)}$&$ 0.857 ^{(2)}$&$ 0.902 ^{(3)}$&$ 1.483^{(4)}$ \\
		
		\hline
	\end{tabular}
	\caption{Test Error Comparisons by MAFE for Different Forests with SUT Trees}\label{real data Test Error MAE (SUT)}
	
\end{table}
The relative MSFE and MAFE are depicted in Figures \ref{real data Relative MSE (SUT)} and \ref{real data Relative MAE (SUT)}, respectively. With SUT trees rather than CART trees, the WRF$_\mathrm{opt}$ performs better at upgrading equal-weight forests. Concerning the MSFE and MAFE, the number of supporting data sets jumps to 11 and 10 out of 12 data sets, respectively. Additionally, the proportion of outperforming rivals climbs to 10 out of 12 data sets for MSFE and 7 out of 12 data sets for MAFE. Notably, the improvement in the EE, SM, and YH data sets are particularly substantial.

Due to the lack of response data for guiding splits, the WRF$_\mathrm{opt}$ with SUT trees has worse predictive power than the WRF$_\mathrm{opt}$ with CART trees. However, it is worthwhile noting that the improvement for forests with equal weights has an essential increase in both the quantity of improvement ratio and the number of supporting data sets. This shows the practical success of our WRF$_\mathrm{opt}$ approaches with weaker base learners.

\section{Conclusion}\label{section:Conclusion}
This paper investigates the weighted RFs for regression. We propose an optimal forest algorithm and its accelerated variant. The proposed methods are asymptotically optimal for the case where structures of trees are independent to the output values in learning samples. Empirical results demonstrate that the proposed methods achieve lower risk compared to RFs with equal weights and other existing unequally weighted forests. While this paper has focused on regression, it would be greatly desirable to study the optimal forests for classification. Another important extension would be to
release the independence of tree architectures from the output values in training data.
\section*{Acknowledgements}
We gratefully acknowledge the support of the NNSFC through grants 12071414 and 11661079.

\newpage

\appendix
\section*{Appendix A. Tree-Building Algorithms}
We will elucidate the differences between the two splitting criteria in this appendix. When constructing RFs using CART trees, we consider Algorithm \ref{algorithm:CART tree}, and when building them with SUT trees, we adopt Algorithm \ref{algorithm:indipendent tree}. The structures of SUT trees are developed in an unsupervised manner, eliminating the reliance on response values during split, whereas CART trees use the information of $\mathbf{y}$ to obtain the best splitting variables and cut points.
They are the same in other procedures, such as growing on the bootstrapped data.

When selecting the probability sequence $\mathcal{P}$ in Algorithm \ref{algorithm:indipendent tree}, we built conventional RFs with CART trees in the validation data to compute variables importance. The variable importance is the total decrease in node impurities from splitting on the variable, averaged over all trees. For regression, the node impurity is measured by residual sum of squares. After that, the probability sequence $\mathcal{P}$ was determined by the normalized variables importance.

\begin{algorithm}[htbp]
	\setstretch{1.3}
	\caption{CART}
	\label{algorithm:CART tree}
	
	\textbf{Split\_a\_node}$(S)$\\
	\KwIn{The local learning subset $S$ corresponding to the node we want to split}
	\KwOut{A split $\left[a<c\right]$ or nothing}
	
	-If \textbf{Stop\_split}$(S)$ is TRUE then return nothing.\\
	
	-Otherwise select $q$ attributes $A_q = \left\{a_{j_1}, \ldots, a_{j_q}\right\}$ randomly among all non constant (in $S$) candidate attributes;\\
	
	-Return the best split $s_*$, where $s_*=$ \textbf{Find\_the\_best\_split}$(S,A_q)$.\\
	
	\hspace*{\fill} \\
	
	\textbf{Find\_the\_best\_split}$(S, A_q)$\\
	\KwIn{The subset $S$ and the selected attribute list $A_q$}
	\KwOut{The best split}
	- Seek the splitting variable $a_j$ and cut point $c$ that solve
	$\min _{j \in \{j_{1},\ldots,j_{q}\}, c} \left\{\min _{c_1} \sum_{\mathbf{x}_i \in \left\{\mathbf{X} \mid \mathbf{X}_j \leq c \right\}}\left(y_i-c_1\right)^2+\min _{c_2} \sum_{\mathbf{x}_i \in \left\{\mathbf{X} \mid \mathbf{X}_j \geq c \right\}}\left(y_i-c_2\right)^2 \right\}$;\\
	- Return the split $\left[a_j<c\right]$.\\
	
	\hspace*{\fill} \\
	
	\textbf{Stop\_split}$(S)$\\
	\KwIn{A subset $S$}
	\KwOut{A boolean}
	- If $|S|<n_{\min }$, then return TRUE;\\
	- If all attributes are constant in $S$, then return TRUE;\\
	- If the output is constant in $S$, then return TRUE;\\
	- Otherwise, return FALSE.
\end{algorithm}

\begin{algorithm}[htbp]
	\setstretch{1.3}
	\caption{SUT}
	\label{algorithm:indipendent tree}
	
	\textbf{Split\_a\_node}$(S)$\\
	\KwIn{The local learning subset $S$ corresponding to the node we want to split}
	\KwOut{A split $\left[a<c \right]$ or nothing}
	
	-If \textbf{Stop\_split}$(S)$ is TRUE then return nothing.\\
	
	-Otherwise select $q$ attributes $\left\{a_1, \ldots, a_q\right\}$ by probability sequence $\mathcal{P}$ among all non constant (in $S$ ) candidate attributes
	\tcp*{Hyper parameter: probability sequence $\mathcal{P}=\{P_1,\cdots,P_p\}$, where $P_j \in [0,1], \forall j=1, \ldots, p$ and $\sum_{j=1}^{p} P_j=1$}
	
	-Draw $q$ splits $\left\{s_1, \ldots, s_K\right\}$, where $s_i=$ \textbf{Pick\_a\_split}$ \left(S, a_i\right), \forall i=1, \ldots, q$;\\
	
	-Return a split $s_*$ such that $\textbf{Score}\left(s_*, S\right)=\max _{i=1, \ldots, q} \textbf{Score}\left(s_i, S\right)$.\\
	
	\hspace*{\fill} \\
	
	\textbf{Pick\_a\_split}$(S, a)$\\
	\KwIn{A subset $S$ and an attribute $a$}
	\KwOut{A split}
	- Let $a_{\max }^S$ and $a_{\min }^S$ be the maximal and minimal value of $a$ in $S$;\\
	- Calculate the cut-point $c \leftarrow (a_{\min }^S, a_{\max }^S)/2$ ;\\
	- Return the split $\left[a<c \right]$.\\
	
	\hspace*{\fill} \\
	
	\textbf{Stop\_split}$(S)$\\
	\KwIn{A subset $S$}
	\KwOut{A boolean}
	- If $|S|<n_{\min }$, then return TRUE.\\
	- If all attributes are constant in $S$, then return TRUE.\\
	- If the output is constant in $S$, then return TRUE.\\
	- Otherwise, return FALSE.
	
	\hspace*{\fill} \\
	
	\textbf{Score}$(s,S)$\\
	\KwIn {A split $s$ and a subset $S$}
	\KwOut {The score of this split method}
	
	-Let $\mathbf{X}_P,\mathbf{X}_L,\mathbf{X}_R$ be the attribute matrix of this local parent node, left daughter, right daughter, respectively;\\
	
	-Let $n_P, n_L, n_R$ be the number of samples contained in the local parent node, left daughter, right daughter, respectively;\\
	
	-Obtain  $\mathbf{\tilde{X}}_P,\mathbf{\tilde{X}}_L,\mathbf{\tilde{X}}_R$ by
	centering and scaling of each column of the matrices $\mathbf{X}_P,\mathbf{X}_L,\mathbf{X}_R$, respectively;\\
	
	-score  $\leftarrow \frac{\|\mathbf{\tilde{X}}_P\| - \frac{n_L}{n_P} \|\mathbf{\tilde{X}}_L\| - \frac{n_R}{n_P} \|\mathbf{\tilde{X}}_R\|}{\|\mathbf{\tilde{X}}_P\|} $;\\
	
	-Return score.
\end{algorithm}

\section*{Appendix B. Detailed Demonstration of Competitors}
\label{wRF_CRF}
In this appendix, we present an exposition of two weighted RF models that have been previously introduced in Section \ref{section:introduction}. Furthermore, we describe a methodology for transforming classification patterns into regression patterns to address predictive regression tasks.

\subsection*{B.1 Weighted RF (wRF)}
Much of the current literature on binary classification pay particular attention to out-of-bag data. Namely, \citet{li2010trees} use the accuracy in the out-of-bag data as an index of the classification ability of a given tree. This metric is subsequently considered to assign weights to the individual trees. \citet{winham2013weighted} provide a family of weights choice based on the prediction error in the out-of-bag data of each tree. The reason why using out-of-bag individuals instead of another shared data set is that it gives internal estimates that are helpful in understanding the predictive performance and how to improve it without testing data set aside
~\citep{breiman2001random}.

Specifically, \citet{winham2013weighted} define the tree-level prediction error, measuring the predictive ability for tree $m$ as follows
\begin{equation}
t \mathrm{PE}_{m}=\frac{1}{\sum_{i=1}^{n} \mathrm{OOB}_{i m}} \sum_{i=1}^{n}\left|v_{ i m}-y_{i}\right| \cdot \mathrm{OOB}_{i m},\label{tPE}
\end{equation}
where $v_{ i m}$ is the vote for subject $i$
in tree $m$ and $\mathrm{OOB}_{i m}$ is the indicator for the out-of-bag status of subject $i$ in tree $m$. By drawing on the concept of $t \mathrm{PE}$, they have been able to show that weights inversely related to $t \mathrm{PE}$ are appropriate. Such as
\begin{equation}
w_{(m)}=1-t\mathrm{PE}_m\label{wRF1},
\end{equation}
\begin{equation}
w_{(m)}=\exp \left( \frac{1}{t\mathrm{PE}_m} \right) \label{wRF2},
\end{equation}
and
\begin{equation}
w_{(m)}={\left(\frac{1}{t\mathrm{PE}_m}\right)}^\lambda \text{for some}~\lambda. \label{wRF3}
\end{equation}
In their proposed wRF algorithm, they normalized weights of the form $$w_{(m)}=\frac{w_{(m)}}{\sum_{m=1}^{M_n}w_{(m)}}.$$

The classification model can be easily turned into a regression model by simply changing (\ref{tPE}) to the following definition
\begin{equation}
t \mathrm{PE}_{m}^{*}=\frac{1}{\sum_{i=1}^{n} \mathrm{OOB}_{i m}} \sum_{i=1}^{n}\left|\hat{f}_{m}(\mathbf{x}_i)-y_{i}\right| \cdot \mathrm{OOB}_{i m}.\label{tPE_star}
\end{equation}
The details of the wRF in regression pattern is in Algorithm \ref{Algorithm wRF}, which selects (\ref{wRF3}) for example. For simplicity, we only present the best result of the wRF family as a representative in Section \ref{section:Real Data Analysis}.

\begin{algorithm}[h]
	\setstretch{1.3}
	\caption{wRF}
	\label{Algorithm wRF}
	\LinesNumbered
	\KwIn {(1) The training data set $\mathcal{D}={\{y_i,\mathbf{x}_i\}}_{i=1}^{n}$
		(2) The number of trees in RF $M_n$ (3) Parameter $\lambda$} 
	\KwOut {Weight vector $\widehat{\mathbf{w}} \in \mathcal{H}$}
	
	\For{m = 1 to $M_n$}{
		Draw a bootstrap data set $\mathcal{D}_{(m)}$ of size $n$ from the training data set $\mathcal{D}$;
		
		Grow a random-forest tree $\hat{f}_{(m)}$ to the bootstrapped data $\mathcal{D}_{(m)}$, by recursively repeating the following steps for each terminal node of
		the tree, until the minimum node size $n_{\min}$ is reached \tcp*{$n_{\min}, q$ are hyper parameters}
		
		\qquad i. Select $q$ variables at random from the $p$ variables;
		
		\qquad ii. Pick the best variable/ split-point among the $q$;
		
		\qquad  iii. Split the node into two daughter nodes.
		
		$t \mathrm{PE}_{m}^{*} \leftarrow \frac{1}{\sum_{i=1}^{n} \mathrm{OOB}_{i m}} \sum_{i=1}^{n}\left|\hat{f}_{(m)}(\mathbf{x_i})-y_{i}\right| \cdot \mathrm{OOB}_{i m}$;
		
		$\widehat{w}_{(m)}\leftarrow {(\frac{1}{t\mathrm{PE}_{m}^{*}})}^\lambda$;	
	}
	$\widehat{\mathbf{w}} \leftarrow{\left(\widehat{w}_{(1)},\ldots,\widehat{w}_{(M_n)}\right)}^\top $;
	
	$\widehat{\mathbf{w}} \leftarrow \frac{\widehat{\mathbf{w}}}{\|\widehat{\mathbf{w}}\|_{1}}$.	
\end{algorithm}

\subsection*{B.2 Ces\'aro RF (CRF) }
Another non-equally weighted RF mentioned earlier is the CRF proposed by~\citet{pham2019cesaro}, which replace the regular average with the Ces\'aro average. Their method is based on a renowned theory that if a sequence converges to a number $c$, then the Ces\'aro sequence also converges to $c$. To implement the CRF, a strategy for sequencing $M_n$ trees from best to worst must be established. This can be done by ranking trees based on their out-of-bag error rates or accuracy on a separate training set. Next, a weight sequence $\left\{w_{(m)}\right\}_{m=1}^{M_n}$ is obtained by arranging weights in descending order, where $w_{(m)}=\sum_{k=m}^{M_n}k^{-1}$, with normalizer being $\sum_{m=1}^{M_n} \sum_{k=m}^{M_n}k^{-1}$.

This classification model can be easily converted into a regression model as well through a simple modification in the sequencing methods. We can draw  $t \mathrm{PE}^{*}$ defined by the wRF algorithm and subsequently rank trees using out-of-bag data. The details of the CRF in regression pattern are in Algorithm \ref{Algorithm CRF}.

\begin{algorithm}[h]
	\setstretch{1.3}
	\caption{CRF}
	\label{Algorithm CRF}
	\LinesNumbered
	\KwIn {(1) The training data set $\mathcal{D}={\{y_i,\mathbf{x}_i\}}_{i=1}^{n}$
		(2) The number of trees in RF $M_n$} 
	\KwOut {Weight vector $\widehat{\mathbf{w}} \in \mathcal{H}$}
	
	\For{m = 1 to $M_n$}{
		Draw a bootstrap data set $\mathcal{D}_{(m)}$ of size $n$ from the training data set $\mathcal{D}$;
		
		Grow a random-forest tree $\hat{f}_{(m)}$ to the bootstrapped data $\mathcal{D}_{(m)}$, by recursively repeating the following steps for each terminal node of
		the tree, until the minimum node size $n_{\min}$ is reached \tcp*{$n_{\min}, q$ are hyper parameters}
		
		\qquad i. Select $q$ variables at random from the $p$ variables;
		
		\qquad ii. Pick the best variable/ split-point among the $q$;
		
		\qquad  iii. Split the node into two daughter nodes.
		
		$t \mathrm{PE}_{m}^{*} \leftarrow \frac{1}{\sum_{i=1}^{n} \mathrm{OOB}_{i m}} \sum_{i=1}^{n}\left|\hat{f}_{(m)}(\mathbf{x_i})-y_{i}\right| \cdot \mathrm{OOB}_{i m}$;
	}
	
	Sequence $\{ t \mathrm{PE}_{1}^{*},\ldots,t \mathrm{PE}_{M_n}^{*} \}$ from smallest to largest;
	
	\For{m = 1 to $M_n$}{
		$r_m \leftarrow$ the order of tree $m$ in sorted sequence;
		
		$\widehat{w}_{(m)}\leftarrow \sum_{k=r_m}^{M_n} \frac{1}{k}$;	
	}	
	$\widehat{\mathbf{w}} \leftarrow{\left(\widehat{w}_{(1)},\ldots,\widehat{w}_{(M_n)}\right)}^\top $;
	
	$\widehat{\mathbf{w}} \leftarrow \frac{\widehat{\mathbf{w}}}{\|\widehat{\mathbf{w}}\|_{1}}$.	
	
\end{algorithm}

\section*{Appendix C. Proofs of Theorems 1 and 2}
In this appendix, we provide  the technical conditions for establishing the asymptotic optimality and adopt them to prove Theorems \ref{theorem OWRF} and \ref{theorem 2step-OWRF}.

\subsection*{C.1 Regularity Conditions}\label{appendix:conditions}

\begin{condition}\label{condition 3}
	$\xi_{n}^{-1} M_{n}^{2}=o(1)$ almost surely.
\end{condition}

\begin{condition}\label{condition 4}
	There exists a positive constant $v$ such that
	$
	\mathbb{E}\left(e_{i}^{4} \mid \mathbf{X}_{i}\right) \leq v<\infty
	$ almost surely for $i=1,\ldots,n$.
\end{condition}

\begin{condition}\label{condition 5}
	$\left\{\min _{1 \leq m \leq M_{n}} \min _{1 \leq l \leq \ell_{(m)}} n_{(m), l}\right\}^{-1} n^{1 / 2}=O(1)$ almost surely.
\end{condition}

\begin{condition}\label{condition extra}
	${\xi}_{n}^{-1} M_{n} n^{1/2} =o(1)$ almost surely.
\end{condition}

Conditions \ref{condition 3} and \ref{condition extra} restrict the increasing rates of the number of trees $M_n$ and the minimum averaging risk $\xi_{n}$. Similar conditions have been considered and discussed by \citet{Zhang2019}, \citet{zhang2021new}, \citet{zou2022optimal}, and others.  Intuitively, these two conditions mean that all trees are misspecified, ruling out the situation where any trees within the RF yield perfect predictions and dominate others.
Condition \ref{condition 4} establishes the boundedness of the conditional moments, which is a mild condition and can be found in much literature.
Condition \ref{condition 5} is a high-level assumption that restricts the structure of the RF and its constituent trees. Specifically, Condition \ref{condition 5} requires that the minimum number of samples in all leaves and all trees should not be of smaller order than $n^{1/2}$, which means that the number of tree leaves has smaller order than $n^{1/2}$. In other words, trees should not be fully developed. Consider the bias-variance decomposition equation
$$
\operatorname{Err}=\operatorname{Bias}^{2}+\mathrm{Var}+\epsilon,
$$
where $\operatorname{Bias}$ is the error produced by the fitted model when it is not capable of representing the true function, $\mathrm{Var}$ is the error resulting from the sampled data, and $\epsilon$ is the error.
Shallow trees have low variances since they are robust to changes in a subset of the sample data. In the meantime, they have high biases because trees are underfitted. In contrast, fully grown trees have low biases and high variances. Therefore, it is advisable to build moderately developed trees in a RF, which are neither too shallow nor too deep. Thus, Condition \ref{condition 5} is reasonable and easy to be satisfied in practice. 

\subsection*{C.2 Preliminary Results}
The following preliminary results will be used in the proofs of Theorems 1 and 2.
Inspired by~\citet{qiu2020mallows}, let $\mathbf{P}_{\mathrm{RT}(m), i\cdot}^{\top}$ be the $i^{\text {th }}$ row of $\mathbf{P}_{\mathrm{RT}(m)}$ for the $m^{\mathrm{th}}$ RT estimator, and $\mathbf{P}_{\mathrm{RT}(m), i, j}$ is the $j^{t h}$ component of $\mathbf{P}_{\mathrm{RT}(m), i\cdot}$. From the discussion in Section \ref{section:Model and Problem Formulation}, we know that the components of $\mathbf{P}_{\mathrm{RT}(m), i \cdot}$ are $\mathbf{1}\left(\mathbf{X}, \mathbf{X}_{i \cdot}\right) / n_{(m), l_i}$, and $\sum_{j=1}^{n} \mathbf{P}_{\mathrm{RT}(m), i, j}=1$, where $n_{(m), l_i}$ is the number of observations in the leaf $l_i$ containing $(y_i,\mathbf{X}_{i\cdot})$, and $\mathbf{1}\left(\mathbf{X}, \mathbf{X}_{i \cdot}\right)$ is an indicator function determined by the relationship between $\mathbf{X}_{i\cdot}$ and $\mathbf{X}$. We reorder $y_i^{\prime}$s and ${\mathbf{X}_{i \cdot}}^{\prime}$s such that the $l^{\mathrm{th}}$ leaf contain observations
$$
\left\{y_{n_{(m),1}+\cdots+n_{(m),l-1}+1}, X_{n_{(m),1}+\cdots+n_{(m),l-1}+1}, \ldots, y_{n_{(m),1}+\cdots+n_{(m),l}}, X_{n_{(m),1}+\cdots+n_{(m),l}}\right\}.
$$
Let $\mathbf{\Pi}_{(m), l}$ be an $n_{(m), l} \times n_{(m), l}$ matrix with all elements being ones and $\bm{\mathcal{L}}_{(m), l}$ be an $n_{(m),l}$-dimensional vector with all elements being ones. \citet{qiu2020mallows} showed that
\begin{equation}
\begin{aligned}
\mathbf{P}_{\mathrm{RT}(m)} &=\operatorname{diag}\left(\mathbf{\Pi}_{(m), 1} n_{(m), 1}^{-1}, \ldots, \mathbf{\Pi}_{(m), \ell} n_{(m), \ell}^{-1}\right) \\
&=\operatorname{diag}\left(\bm{\mathcal{L}}_{(m), 1} \bm{\mathcal{L}}_{(m), 1}^{\top} n_{(m), 1}^{-1},\ldots, \bm{\mathcal{L}}_{(m), \ell} \bm{\mathcal{L}}_{(m), \ell}^{\top} n_{(m), \ell}^{-1}\right).
\end{aligned}\label{P_structure}
\end{equation}
Additionally, they have proved that $\mathbf{P}_{\mathrm{RT}(m)} \text{ for }m=1,\cdots,M_n$ satisfies the properties in Lemma \ref{properties in MAML}. For convenience, they assume $\mathbf{X}$ be non-stochastic instead of stochastic in all proofs. Nevertheless, the same conclusions can still be drawn under the assumption of  stochastic $\mathbf{X}$. This can be achieved by substituting expectations with conditional expectations in the equations, and applying the Law of Iterated (or Total) Expectation, Pull-out rule and Lebesgue Dominated Convergence Theorem in
the process of obtaining orders of probabilities. An example of proof in the case of stochastic $\mathbf{X}$ is demonstrated in Appendix C.5.

\begin{lem}[\citealp{qiu2020mallows}] \label{properties in MAML}
	 For each $m \in \{1,\ldots,M_n\}$, $\mathbf{P}_{\mathrm{RT}(m)}$ has the following properties.
	\begin{enumerate}
		\item There exists a positive constant $c_0$ such that for all $m, s \in\left\{1, \ldots, M_{n}\right\}$,
		$$
		\operatorname{trace}\left(\mathbf{P}_{\mathrm{RT}(m)} \mathbf{P}_{\mathrm{RT}(m)}^{\top}\right) \geq c_{0}>0 \quad \text { and } \quad \operatorname{trace}\left(\mathbf{P}_{\mathrm{RT}(m)} \mathbf{P}_{\mathrm{RT}(s)}^{\top}\right) \geq 0
		$$
		almost surely.
		
		\item There exists a positive constant $c_{1}$ such that for all $m \in\left\{1, \ldots, M_{n}\right\}$,
		$$
		\zeta_{\max }\left(\mathbf{P}_{\mathrm{RT}(m)} \mathbf{P}_{\mathrm{RT}(m)}^{\top}\right) \leq c_{1},
		$$
		almost surely, where $\zeta_{\max }(\mathbf{B})$ denotes the largest singular value of a matrix $\mathbf{B}$.
		
		\item There exists a positive constant $c_{2}$ such that for all $m, j \in\left\{1, \ldots, M_{n}\right\}$,
		$$
		\operatorname{trace}\left(\mathbf{P}_{\mathrm{RT}(m)}^{2}\right) \leq c_{2} \operatorname{trace}\left(\mathbf{P}_{\mathrm{RT}(m)}^{\top} \mathbf{P}_{\mathrm{RT}(m)}\right),
		$$
		and
		$$
		\operatorname{trace}\left(\mathbf{P}_{\mathrm{RT}(j)}^{\top} \mathbf{P}_{\mathrm{RT}(m)} \mathbf{P}_{\mathrm{RT}(j)}^{\top} \mathbf{P}_{\mathrm{RT}(m)}\right) \leq c_{2} \operatorname{trace}\left(\mathbf{P}_{\mathrm{RT}(m)}^{\top} \mathbf{P}_{\mathrm{RT}(m)}\right)
		$$
		almost surely.
		
		\item We have
		$\max_{1 \leq m \leq M_{n}} \max_{1 \leq i \leq n} \iota_{ii}^{(m)\mathrm{RT}}=O\left(n^{-1 / 2}\right)$ almost surely, where $\iota_{ii}^{(m)\mathrm{RT}}$ is the $i^{\text {th }}$ diagonal element of $\mathbf{P}_{\mathrm{RT}(m)}$.

            \item For each $m\in \{1,\ldots,M_n\}$, let $\mathbf{A}_{(m)}$ be a random matrix with elements $A_{(m)ij}\in \{0,1\}, A_{(m)ii}=0$ and
    $\mathbb{E}(A_{(m)ij})={(n-1)}^{-1}$ for $i \neq j$. Then, results 1-4 above still hold if substitute $\mathbf{P}_{\mathrm{RT}(m)}$ with $\mathbf{P}_{\mathrm{RT}(m)} \mathbf{A}_{(m)}$.
		
\end{enumerate}
\end{lem}
Next, we introduce two other lemmas for proving Theorems 1 and 2.
\begin{lem}[\citealp{gao2019frequentist}]\label{gao}
	Let
	$$
	\widetilde{\mathbf{w}}=\operatorname{argmin}_{\mathbf{w} \in \mathcal{H}}\left\{L_{n}(\mathbf{w})+a_{n}(\mathbf{w})+b_{n}\right\},
	$$
	where $a_{n}(\mathbf{w})$ is a term related to $\mathbf{w}$ and $b_{n}$ is a term unrelated to $\mathbf{w}$. If
	$$
	\sup _{\mathbf{w} \in \mathcal{H}}\left|a_{n}(\mathbf{w})\right| / R_{n}(\mathbf{w})=o_{p}(1), \sup _{\mathbf{w} \in \mathcal{H}}\left|R_{n}(\mathbf{w})-L_{n}(\mathbf{w})\right| / R_{n}(\mathbf{w})=o_{p}(1),
	$$
	and there exists a constant $c$ and a positive integer $n^{*}$ so that when $n \geq n^{*}$,  $inf_{\mathbf{w} \in \mathcal{H}} R_{n}(\mathbf{w}) \geq$ $c>0$ almost surely, then $L_{n}(\widetilde{\mathbf{w}}) / \inf_{\mathbf{w} \in \mathcal{H}} L_{n}(\mathbf{w}) \rightarrow 1$ in probability.
\end{lem}

\begin{lem}[\citealp{saniuk1987matrix}] \label{inequality of trace}
For any $n \times n$ matrices $\mathbf{G}_1$ and $\mathbf{G}_2$ with both $\mathbf{G}_1, \mathbf{G}_2 \geqslant 0$,
$$
\operatorname{trace}(\mathbf{G}_1 \mathbf{G}_2) \leqslant \| \mathbf{G}_1\|_2 \operatorname{trace}(\mathbf{G}_2),
$$
where $\| \cdot \|_2$ denotes the spectral norm or largest singular value.
\end{lem}

\subsection*{C.3 Proof of Theorem \ref{theorem OWRF}} \label{subsection: proof of theorem 1}
It remains to verify that Conditions \ref{condition 3} to \ref{condition 5} can guarantee Conditions C.4 - C.9 in~\citet{qiu2020mallows}.
Just note that the base learners in a RF employ two extra techniques than the  regression trees. They are built on data sampled randomly with replacement and split nodes on a random subset of features, which are also employed in the bootstrap aggregating algorithm and the conventional RF algorithm, respectively. To get $\mathbf{P}_{\mathrm{BL}{(m)}}$ from $\mathbf{P}_{\mathrm{RT}{(m)}}$ for $m \in \left\{1, \ldots, M_n\right\}$, it is sufficient to integrate these two randomization techniques into the formulation of $\mathbf{P}_{\mathrm{RT}{(m)}}$.

Let $\mathbf{P}_{\mathrm{RT}}$ be an $n \times n$ matrix, of which the $i^{\mathrm{th }}$ row is $\mathbf{P}_{\mathrm{RT}}^{\top}(\mathbf{x}_{i},\mathbf{X},\mathbf{y},\mathbf{\Theta}^{\mathrm{RT}})$. The random vector $\mathbf{\Theta}^{\mathrm{RT}}$ characterizes the randomness in the regression tree, such as how to split nodes. Denote $\mathbf{\Theta}^{\mathrm{RT}}$ when the best split is found over a randomly selected subset of $q$ predictor variables instead of all $p$ predictors by $\tilde{\mathbf{\Theta}}$. Let $\tilde{\mathbf{P}}$ be an $n \times n$ matrix, of which the $i^{\mathrm{th }}$ row is $\mathbf{P}_{\mathrm{RT}}^{\top}(\mathbf{x}_{i},\mathbf{X},\mathbf{y},\tilde{\mathbf{\Theta}})$. For each $m \in \{1,\ldots,M_n\}$, let $\tilde{\mathbf{P}}_{(m)}$ denote the $\tilde{\mathbf{P}}$ related to the $m^{\mathrm{th}}$ tree, which remains a diagonal block matrix after data reordering, and has the same results stated in Lemma \ref{properties in MAML} as $\mathbf{P}_{\mathrm{RT}(m)}$.

Inspired by~\citet{qiu2020mallows}, we exploit selection matrices to represent the bootstrapping procedure.
Let $\mathbf{A}$ be a random matrix with $A_{ij}\in \{0,1\}, A_{ii}=0$, such that
$\mathbb{E}(A_{ij})={(n-1)}^{-1}$ for $i \neq j$. Consequently, we have
\begin{equation}
\mathbf{P}_{\mathrm{BL}{(m)}}=\tilde{\mathbf{P}}_{(m)} \mathbf{A}_{(m)},\label{s2}
\end{equation}
where $\mathbf{A}_{(m)}$ is the selection matrix $\mathbf{A}$ relating the $m^{\mathrm{th}}$ bootstrapping procedure.
Note that for each $m \in \{1,\ldots,M_n\}$, we have
$$
\{ \max_{1 \leq m \leq M_{n}} \max_{1 \leq i \leq n} \iota_{ii}^{(m)\mathrm{BL}}\}
{\{\max_{1 \leq m \leq M_{n}} \max_{1 \leq i \leq n} \iota_{ii}^{(m)\mathrm{RT}} \}}^{-1}=O(1),
$$	
almost surely,
where $\iota_{ii}^{(m)\mathrm{BL}}$ is the $i^{\text {th }}$ diagonal element of $\mathbf{P}_{\mathrm{BL}(m)}$.
Then, $\mathbf{P}_{\mathrm{BL}{(m)}}$ inherits properties 1-4 as described in  Lemma \ref{properties in MAML}, with the aid of result 5 from the same lemma.
Further, Conditions \ref{condition 3} and \ref{condition 4} in this paper are analogous to Conditions C.7 and C.8 in \citet{qiu2020mallows}, respectively.

Based on the above results, Conditions \ref{condition 3} - \ref{condition 5} can deduce technical Conditions C.4 - C.9 showed by~\citet{qiu2020mallows}, which implies that the asymptotic optimality in the sense
of achieving the lowest possible squared error can be established.

By (\ref{asymptotic optimality:Ln'}) in Section \ref{section:Mallows-type Averaging Random Forest}, and (A34) in Proof of Theorem 2 by \citet{qiu2020mallows} , we further have
\begin{equation}
L_n(\widehat{\mathbf{w}}) \xi_{n}^{-1}=1+o_p(1).\label{uniform inte c6}
\end{equation}
This is because
\begin{align}
	L_{n}(\widehat{\mathbf{w}}) \xi_{n}^{-1} \leq & \sup _{\mathbf{w} \in \mathcal{H}}\left\{L_{n}(\widehat{\mathbf{w}}) L_{n}^{-1}(\mathbf{w})\right\} \sup _{\mathbf{w} \in \mathcal{H}}\left\{L_{n}(\mathbf{w}) R_{n}^{-1}(\mathbf{w})\right\} \nonumber\\
	\leq & \sup _{\mathbf{w} \in \mathcal{H}}\left\{L_{n}(\widehat{\mathbf{w}}) L_{n}^{-1}(\mathbf{w})\right\} \times\left[1+\sup _{\mathbf{w} \in \mathcal{H}}\left\{\left|L_{n}(\mathbf{w})-R_{n}(\mathbf{w})\right| R_{n}^{-1}(\mathbf{w})\right\}\right] \nonumber\\
	=& 1+o_{p}(1),
\end{align}\label{ineq1}
and
\begin{align}
L_{n}(\widehat{\mathbf{w}}) \xi_{n}^{-1} &\geq \sup _{\mathbf{w} \in \mathcal{H}}\left\{L_{n}(\widehat{\mathbf{w}}) L_{n}^{-1}(\mathbf{w})\right\} \inf _{\mathbf{w} \in \mathcal{H}}\left\{L_{n}(\mathbf{w}) R_{n}^{-1}(\mathbf{w})\right\} \nonumber\\
&=\sup _{\mathbf{w} \in \mathcal{H}}\left\{L_{n}(\widehat{\mathbf{w}}) L_{n}^{-1}(\mathbf{w})\right\} \times\left(1+\inf _{\mathbf{w} \in \mathcal{H}}\left[\left\{L_{n}(\mathbf{w})-R_{n}(\mathbf{w})\right\} R_{n}^{-1}(\mathbf{w})\right]\right) \nonumber\\
&\geq \sup _{\mathbf{w} \in \mathcal{H}}\left\{L_{n}(\widehat{\mathbf{w}}) L_{n}^{-1}(\mathbf{w})\right\}
\times\left[1-\sup _{\mathbf{w} \in \mathcal{H}}\left\{\left|L_{n}(\mathbf{w})-R_{n}(\mathbf{w})\right| R_{n}^{-1}(\mathbf{w})\right\}\right] \nonumber\\
&=1+o_{p}(1) ,
\end{align}\label{ineq2}
which is identical with the part in Appendix A.4 by~\citet{zhang2019parsimonious}.
Consequently, from the uniform integrability of $\left\{L_n(\widehat{\mathbf{w}})-\xi_n\right\}\xi_{n}^{-1}$, it is apparent that $\mathbb{E}\left[\left\{L_n(\widehat{\mathbf{w}})-\xi_n\right\}\xi_{n}^{-1}\right] \rightarrow 0$, by which we obtain (\ref{risk}).

\subsection*{C.4 Proof of Theorem \ref{theorem 2step-OWRF}}
Based on Lemma \ref{gao}, we now present the proof of Theorem \ref{theorem 2step-OWRF}.
It is seen that
$$C_{n}^{\prime \prime}(\mathbf{w})=C_{n}(\mathbf{w})+2 \sum_{i=1}^{n} \left(\tilde{e}_{i}^{2}-e_{i}^{2}\right) P_{ii}(\mathbf{w}).$$
Hence, from Lemma \ref{gao} above and (24) in \citet{qiu2020mallows}, in order to prove (\ref{asymptotic optimality:Ln''}), we need only to verify that
\begin{equation}
\sup _{\mathbf{w} \in \mathcal{H}}\left\{\left|\sum_{i=1}^{n}\left(\tilde{e}_{i}^{2}-e_{i}^{2}\right) P_{ii}(\mathbf{w})\right| / R_{n}(\mathbf{w})\right\}=o_{p}(1).
\end{equation}
Let $\iota_{i i}^{(m)}$ be the $i^{\text {th }}$ diagonal element of $\mathbf{P}_{\mathrm{B L}{(m)}},  \mathbf{Q}_{(m)}=\operatorname{diag}\left(\iota_{11}^{(m)}, \cdots, \iota_{n n}^{(m)}\right)$, $\mathbf{Q}(\mathbf{w})=\sum_{m=1}^{M_{n}} w_{(m)} \mathbf{Q}_{(m)}$, and $\mathbf{K}_n=\operatorname{diag}\left(\tilde{e}_{1}^{2}-e_{1}^{2}, \cdots, \tilde{e}_{n}^{2}-e_{n}^{2}\right)$. Then, we have
$$
\quad \sup _{\mathbf{w} \in \mathcal{H}}\left\{\left|\sum_{i=1}^{n}\left(\tilde{e}_{i}^{2}-e_{i}^{2}\right) P_{ii}(\mathbf{w})\right| / R_{n}(\mathbf{w})\right\}
=
\sup_{\mathbf{w} \in \mathcal{H}} \frac{\left| \operatorname{trace}\left\{\mathbf{Q}(\mathbf{w}) \mathbf{K}_n \right\}\right|} {R_{n}(\mathbf{w})}.
$$
We observe that for any $\delta>0$,
$$
\begin{aligned}
&\operatorname{Pr}\left(\sup_{\mathbf{w} \in \mathcal{H}}\left|\frac{\operatorname{trace}\left\{\mathbf{Q}(\mathbf{w}) \mathbf{K}_n\right\}}{R_n(\mathbf{w})}\right|>\delta\right) \\
\leqslant  &\sum_{m=1}^{M_n} \operatorname{Pr}\left(\left|\frac{\operatorname{trace}\left(\mathbf{Q}_{(m)} \mathbf{K}_n\right)}{R_n(\mathbf{w})}\right|>\delta\right) \\
\leqslant & {\delta}^{-1} \sum_{m=1}^{M_n} \mathbb{E}\left\{\left|\frac{\operatorname{trace}\left(\mathbf{Q}_{(m)} \mathbf{K}_n\right)}{R_n(\mathbf{w})}\right|\right\} \\
\leqslant &{\delta}^{-1} \sum_{m=1}^{M_n} \mathbb{E}\left\{\left|\operatorname{trace}\left(\mathbf{Q}_{(m)} \mathbf{K}_n\right)\right| \xi_n^{-1}\right\} \\
\leqslant & {\delta}^{-1} \sum_{m=1}^{M_n} \mathbb{E}\left\{\operatorname{trace}\left(\mathbf{Q}_{(m)}\right)\left\|\mathbf{K}_n\right\|_{2} \xi_n^{-1}\right\} \\
= & {\delta}^{-1} \sum_{m=1}^{M_n} \mathbb{E}\left(\ell_{(m)} \left\|\mathbf{K}_n\right\|_{2} \xi_n^{-1}\right) \\
\leqslant & c_1 {\delta}^{-1} M_n \mathbb{E}\left(n^{1 / 2}\left\|\mathbf{K}_n\right\|_{2} \xi_n^{-1}\right) \\
\leqslant & c_1 {\delta}^{-1} M_n \mathbb{E}^{1 / 2}\left(n^{1 / 2} \xi_n^{-1}\right)^2 \mathbb{E}^{1 / 2}\left\|\mathbf{K}_n\right\|_{2}^2 \\
\leqslant &c_{2}{\delta}^{-1} \mathbb{E}^{1 / 2}{\left(M_{n} n^{1/2} \xi_n^{-1}\right)}^{2},
\end{aligned}
$$
where $c_1$ and $c_2$ are positive constants, the second inequality follows from the Chebyshev's Inequality, the fourth inequality is obtained by Lemma \ref{inequality of trace}, the fifth inequality comes from Condition \ref{condition 5}, and the last second inequality is from the Cauchy-Schwarz Inequality. Thus, (\ref{asymptotic optimality:Ln''}) is proved by Condition \ref{condition extra} and the Lebesgue's Dominated Convergence Theorem.
Similar to the proof techniques of Theorem \ref{theorem OWRF}, we have $
L_n(\widetilde{\mathbf{w}}) \xi_{n}^{-1}=1+o_p(1)$. This completes the proof of (\ref{risk''}).

\subsection*{C.5 Verifying Results of \citet{qiu2020mallows} for Stochastic $\mathbf{X}$}
For the sake of convenience, \citet{qiu2020mallows} assume in all proofs that $\mathbf{X}$ is non-stochastic rather than stochastic. However, theorems and corollaries in \citet{qiu2020mallows} still hold when $\mathbf{X}$ is assumed to be stochastic. We take
$$
\sup _{\mathbf{w} \in \mathcal{H}} \left|\mathbf{e}^{\top} \mathbf{A}(\mathbf{w}) \boldsymbol{\mu} \right| / {R_{n}(\mathbf{w})}=o_p(1)
$$
for example, which is labeled as (A6) in Appendix A.2 of \citet{qiu2020mallows}. \\
\textbf{Proof of (A6) in \citet{qiu2020mallows} when $\mathbf{X}$ is Stochastic}. Let $\mathbf{\Phi}=\left({\boldsymbol{\mu}}^{\top} \mathbf{A}_{(m)}^{\top} \mathbf{A}_{(s)} \boldsymbol{\mu}\right)_{M_{n} \times M_{n}}$, i.e., the ${m s}^{\text {th }}$ component of $\mathbf{\Phi}$ is $\boldsymbol{\mu}^{\top} \mathbf{A}_{(m)}^{\top} \mathbf{A}_{(s)} \boldsymbol{\mu}$, $\mathbf{G}_{n \times M_{n}}=\left(\mathbf{A}_{(1)} \boldsymbol{\mu}, \ldots, \mathbf{A}_{\left(M_{n}\right)} \boldsymbol{\mu}\right), \mathbf{\Psi}=\left\{\sigma^{2} \operatorname{trace}\left(\mathbf{P}_{(m)} \mathbf{P}_{(s)}^{\top}\right)\right\}_{M_{n} \times M_{n}}$, and
$$
\mathbf{\Psi}_{0}=\sigma^{2} \operatorname{diag}\left\{\operatorname{trace}\left(\mathbf{P}_{(1)} \mathbf{P}_{(1)}^{\top}\right), \ldots, \operatorname{trace}\left(\mathbf{P}_{\left(M_{n}\right)} \mathbf{P}_{\left(M_{n}\right)}^{\top}\right)\right\}.
$$
So $\mathbf{\Phi}=\mathbf{G}^{\top} \mathbf{G}$. For any $\mathbf{w} \in \mathcal{H}$,
\begin{equation}
    \mathbf{w}^{\top} \mathbf{\Psi}_{0} \mathbf{w} \leqslant \mathbf{w}^{\top} \mathbf{\Psi} \mathbf{w},\label{A10}
\end{equation}
because for any $m, s \in\left\{1, \ldots, M_{n}\right\}, \mathbf{w}_{(m)} \geq 0, \mathbf{w}_{(s)} \geq 0$ and trace $\left(\mathbf{P}_{(m)} \mathbf{P}_{(s)}^{\top}\right) \geq 0$ by Condition C.4. In addition,
\begin{align}
R_{n}(\mathbf{w}) & =\mathbb{E}\left\{ \left.\|\mathbf{P}(\mathbf{w}) \boldsymbol{\mu}-\boldsymbol{\mu}+\mathbf{P}(\mathbf{w}) \mathbf{e}\|^{2} \right| \mathbf{X}\right\} \nonumber\\
& =\|\mathbf{A}(\mathbf{w}) \boldsymbol{\mu}\|^{2}+\sigma^{2} \operatorname{trace}\left\{\mathbf{P}(\mathbf{w}) \mathbf{P}(\mathbf{w})^{\top}\right\} \nonumber\\
& =\mathbf{w}^{\top}(\mathbf{\Phi}+\mathbf{\Psi}) \mathbf{w} \nonumber\\
& \geqslant \mathbf{w}^{\top}\left(\mathbf{\Phi}+\mathbf{\Psi}_{0}\right) \mathbf{w},\label{A11}
\end{align}
where the last step is from (\ref{A10}). We also have
\begin{equation}
 \mathbf{\Phi}+\mathbf{\Psi}_{0}>0,\label{A12}
\end{equation}
because $\mathbf{\Phi}=\mathbf{G}^{\top} \mathbf{G}$ and $\mathbf{\Psi}_{0}>0$ by Condition C.4 in \citet{qiu2020mallows}.

Let $\boldsymbol{\rho}=\left(\mathbf{e}^{\top} \mathbf{A}_{(1)} \boldsymbol{\mu} \ldots, \mathbf{e}^{\top} \mathbf{A}_{\left(M_{n}\right)} \boldsymbol{\mu}\right)^{\top}$. It is straightforward to show that
\begin{equation}
 \mathbb{E}(\boldsymbol{\rho} | \mathbf{X})=0,\label{A13}
\end{equation}
and
\begin{equation}
   \operatorname{Var}(\boldsymbol{\rho} | \mathbf{X})=\mathbb{E}\left(\left. \boldsymbol{\rho}\boldsymbol{\rho} ^{\top} \right| \mathbf{X}\right)=\mathbb{E} \left. \left\{\left(\mathbf{e}^{\top} \mathbf{A}_{(m)} \boldsymbol{\mu} \boldsymbol{\mu}^{\top} \mathbf{A}_{(s)} \mathbf{e}\right)_{M_{n} \times M_{n}}  \right| \mathbf{X}\right\}=\sigma^{2} \mathbf{\Phi}. \label{A14}
\end{equation}
It is seen that
\begin{align}
\sup _{\mathbf{w} \in \mathcal{H}} \frac{\left\{\mathbf{e}^{\top} \mathbf{A}(\mathbf{w}) \boldsymbol{\mu}\right\}^{2}}{R_{n}^{2}(\mathbf{w})} & =\sup _{\mathbf{w} \in \mathcal{H}} \frac{\left(\sum_{m=1}^{M_{n}} \mathbf{w}_{(m)} \mathbf{e}^{\top} \mathbf{A}_{(m)} \boldsymbol{\mu}\right)^{2}}{R_{n}^{2}(\mathbf{w})} \nonumber\\
& =\sup _{\mathbf{w} \in \mathcal{H}} \frac{\left(\mathbf{w}^{\top}
\boldsymbol{\rho}\right)^{2}}{R_{n}^{2}(\mathbf{w})} \nonumber\\
& \leqslant \sup _{\mathbf{w} \in \mathcal{H}} \frac{\left(\mathbf{w}^{\top} \boldsymbol{\rho}\right)^{2}}{\mathbf{w}^{\top}\left(\mathbf{\Phi}+\mathbf{\Psi}_{0}\right) \mathbf{w}} \sup _{\mathbf{w} \in \mathcal{H}} \frac{1}{R_{n}(\mathbf{w})} \nonumber\\
& \leqslant \xi_{n}^{-1} \boldsymbol{\rho}^{\top}\left(\mathbf{\Phi}+\mathbf{\Psi}_{0}\right)^{-1} \boldsymbol{\rho},\label{A15}
\end{align}
where the third step is from (\ref{A11}) and the last step is from (\ref{A12}) and Lemma 1 in \citet{qiu2020mallows}. By Markov Inequality, we have that for any $\delta>0$,
\begin{align}
& \operatorname{Pr}\left\{\xi_{n}^{-1} \boldsymbol{\rho}^{\top}\left(\mathbf{\Phi}+\mathbf{\Psi}_{0}\right)^{-1} \boldsymbol{\rho}>\delta\right\} \nonumber \\
\leqslant & \delta^{-1}  \mathbb{E}\left\{\xi_{n}^{-1}\boldsymbol{\rho}^{\top}\left(\mathbf{\Phi}+\mathbf{\Psi}_{0}\right)^{-1} \boldsymbol{\rho}\right\} \nonumber\\
=& \delta^{-1}  \mathbb{E} \left\{ \mathbb{E}\left[\left. \xi_{n}^{-1}\boldsymbol{\rho}^{\top}\left(\mathbf{\Phi}+\mathbf{\Psi}_{0}\right)^{-1} \boldsymbol{\rho} \right| \mathbf{X}\right]\right\} \nonumber\\
=& \delta^{-1}  \mathbb{E} \left[ \xi_{n}^{-1}\mathbb{E}\left\{ \left. \boldsymbol{\rho}^{\top}\left(\mathbf{\Phi}+\mathbf{\Psi}_{0}\right)^{-1} \boldsymbol{\rho} \right| \mathbf{X}\right\} \right]\nonumber\\
= & \delta^{-1}  \mathbb{E} \left[\xi_{n}^{-1} \sigma^{2} \operatorname{trace}\left\{\left(\mathbf{\Phi}+\mathbf{\Psi}_{0}\right)^{-1} \mathbf{\Phi}\right\} \right] \nonumber\\
\leqslant & \delta^{-1}  \mathbb{E} \left[\xi_{n}^{-1} \sigma^{2} \operatorname{trace}\left\{\left(\mathbf{\Phi}+\mathbf{\Psi}_{0}\right)^{-1} \mathbf{\Phi}+\mathbf{\Psi}_{0}^{1 / 2}\left(\mathbf{\Phi}+\mathbf{\Psi}_{0}\right)^{-1} \mathbf{\Psi}_{0}^{1 / 2}\right\} \right] \nonumber \\
= & {\delta}^{-1}  \mathbb{E} \left( \xi_{n}^{-1} \sigma^{2} M_{n} \right), \label{A16}
\end{align}
where the second step follows from the Law of Iterated (or Total) Expectation, the third step is obtained by the Pull-out rule, the fourth step is guaranteed by (\ref{A13}) and (\ref{A14}). Combining (\ref{A15}), (\ref{A16}) and Condition \ref{condition 3}, we obtain (A6) in \citet{qiu2020mallows} by the Lebesgue Dominated Convergence Theorem. $\hfill\blacksquare$

This demonstration bears a striking resemblance to Proof of (A6) in Appendix A.2 of \citet{qiu2020mallows}. The only modification lies in the substitution of expectations with conditional expectations in (\ref{A13}) and (\ref{A14}), and the utilization of the Law of Iterated Expectation, Pull-out rule, and Lebesgue Dominated Convergence Theorem in (\ref{A16}). Equations (A7) - (A9) and (A34) - (A35) for proving Theorems 1 and 2 in \citet{qiu2020mallows} can also be extrapolated using the same techniques.

\vskip 0.2in
\bibliography{sample}

\end{document}